\documentclass{article}

\PassOptionsToPackage{numbers,sort&compress}{natbib}
\usepackage[preprint]{neurips_2026}

\usepackage[utf8]{inputenc}
\usepackage[T1]{fontenc}
\usepackage{hyperref}
\usepackage{url}
\usepackage{booktabs}
\usepackage{amsfonts}
\usepackage{amsmath}
\usepackage{amssymb}
\usepackage{enumitem}
\usepackage{nicefrac}
\usepackage{microtype}
\usepackage{xcolor}
\usepackage{multirow}
\usepackage{graphicx}
\usepackage{tikz}
\usetikzlibrary{arrows.meta,positioning,calc,decorations.pathreplacing,fit,backgrounds}
\usepackage{wrapfig}
\usepackage{float}
\usepackage{caption}
\usepackage[framemethod=tikz]{mdframed}
\newcommand{\promptheader}[1]{%
  \noindent
  \colorbox{black}{\parbox{\dimexpr\linewidth-2\fboxsep\relax}{%
    \color{white}\bfseries #1}}%
  \nopagebreak\par\vspace{-1pt}}
\newmdenv[
  linecolor=black, linewidth=0.5pt,
  roundcorner=4pt,
  innerleftmargin=8pt, innerrightmargin=8pt,
  innertopmargin=4pt, innerbottommargin=4pt,
  skipabove=10pt, skipbelow=10pt
]{promptframe}
\usepackage{amsthm}

\theoremstyle{remark}

\newcommand{\ours}{LLM-GNN Co-Teaching}
\newcommand{\best}[1]{\textbf{#1}}

\title{Beyond the Golden Teacher: Enhancing Graph Learning through LLM-GNN Co-teaching}

\author{%
  Zhuoyi Peng$^{1}$ \quad Hanlin Gu$^{2}$ \quad Lixin Fan$^{2}$ \quad Yi Yang$^{1}$ \\[0.4em]
  $^{1}$The Hong Kong University of Science and Technology \quad $^{2}$WeBank
}

\begin{document}

\maketitle

\begin{abstract}
Text-attributed graphs (TAGs) underlie real-world applications such as citation networks, social media, and e-commerce. Few-shot graph learning on TAGs is hard: with only a handful of labels per class and the rest of the graph unannotated, neither GNNs nor LLMs can learn well on their own. GNNs read topology and fail on cold nodes; LLMs read text and fail on text-ambiguous nodes. Existing LLM-GNN methods all follow the same recipe: \emph{designate one model as the golden teacher and use its outputs (e.g., features or pseudo-labels) to supervise the other.} We argue this golden-teacher assumption breaks under sparse supervision: neither model is golden, and treating either as such transfers its blind spots into the student. We therefore ask: \emph{can we avoid designating either model as the golden teacher, and still perform effective graph learning?} We answer with \ours{}, a bidirectional co-teaching framework in which neither model is fixed as teacher. The GNN and LLM exchange their most confident pseudo-labels under an architecture-specific small-loss criterion, and both update every round. Supervision is then mined from the trajectory: whenever a node moves from cross-model contradiction at round $t$ to cross-model agreement at round $t+1$, the LLM's two answers on the same input form a preference pair (\emph{old contradicting self} $\prec$ \emph{new peer-endorsed self}) for DPO training. We call this Round-based Pseudo-Label Preference Optimization (RPL-PO). On six benchmarks, \ours{} consistently outperforms GNN-as-Judge and all prior methods, with absolute 3-shot gains of 7.86\% on Cora and 7.73\% on ogbn-arxiv; improvements carry over to 5-shot and to zero-shot cross-dataset transfer. Error-structure analysis further shows that abandoning the golden-teacher assumption substantially improves the LLM's graph learning capability on challenging samples. Code: \url{https://github.com/llmgnncoteaching/LLM-GNN-Coteaching}.
\end{abstract}

\section{Introduction}
\label{sec:intro}

Text-attributed graphs (TAGs)~\citep{hu2020open,yang2016revisiting,sen2008collective,he2023harnessing} underlie real-world applications such as citation networks, social media, recommendation, and e-commerce, where each node carries raw text alongside graph topology. The rise of Large Language Models (LLMs)~\citep{brown2020language,achiam2023gpt,grattafiori2024llama} has driven growing interest in using them for TAG learning~\citep{chen2023label,he2023harnessing,tang2024graphgpt,ye2024language,chen2024llaga,liu2023one}. Most existing work on TAG learning, however, focuses on the supervised setting where abundant labels are available and both models can be reliably fine-tuned~\citep{he2023harnessing,zhao2022learning,tang2024graphgpt,chen2024llaga,ye2024language,wang2024instructgraph,liu2023one}. Real-world TAGs are rarely labeled at this scale: only a handful of labels per class are typically available, and the bulk of the graph carries no supervision~\citep{ding2020graph,ding2022meta,wang2022graph,huang2020graph,sun2020multi}. Under this few-shot regime, neither GNNs~\citep{kipf2016semi,wu2019simplifying,gasteiger2018predict,xu2018representation} nor LLMs work well alone: GNNs read topology and fail on cold (low-degree) nodes whose neighborhoods provide too little signal~\citep{li2018deeper,zhu2020beyond}, while LLMs read text and fail when text is short or class-ambiguous~\citep{huang2023can,wu2025llms,dai2024large,guo2023gpt4graph}. Their disjoint failure modes have motivated a substantial line of work combining them.

Existing LLM-GNN methods all share a common structure: one model is designated as a fixed teacher whose outputs are treated as ground truth, and the other is trained to match those outputs. We refer to this shared structural assumption as the \emph{golden-teacher assumption}. Prior approaches differ only in which side is designated as golden. \emph{LLM-as-Enhancers}~\citep{he2023harnessing,zhao2022learning,li2023grenade,yang2021graphformers} freeze LLM-derived features or explanations and train a downstream GNN to imitate them. \emph{LLM-as-Predictor} methods~\citep{tang2024graphgpt,chen2024llaga,ye2024language,liu2023one,wang2024instructgraph,wang2024llms,hu2024let} treat the once-instruction-tuned LLM as the golden predictor, typically prompting it with structural tokens. \emph{GNN-as-Judge}~\citep{xu2026gnn} reverses the direction: a once-trained GNN's verdicts filter or re-weight pseudo-labels for LLM fine-tuning. In every case, supervision flows in one direction from a fixed teacher, and the student has no way to revise what the teacher said.

The golden-teacher assumption breaks under sparse supervision. \textbf{With few labels per class, neither model is reliable enough to serve as the golden teacher: the GNN cannot learn good representations for cold nodes, and the LLM cannot disambiguate short or class-ambiguous text without exemplars.} Treating either as golden transfers its blind spots into the student wholesale, and unidirectional supervision leaves the student no way to revise what the teacher said. The question that few-shot LLM-GNN learning has not yet asked is therefore:

\begin{quote}
\emph{Can we avoid designating either model as the golden teacher, and still perform effective graph learning?}
\end{quote}

The question is not trivial: with only a few labeled anchors as direct supervision, two weak models updating each other freely can collapse onto each other's mistakes rather than converge toward truth. The framework needs a mechanism that extracts reliable supervisory signal from their joint dynamics.

\definecolor{gnnblue}{HTML}{4472C4}
\definecolor{llmorange}{HTML}{ED7D31}
\definecolor{leftbg}{HTML}{FBE6E6}
\definecolor{rightbg}{HTML}{EAF5EE}
\definecolor{prefbg}{HTML}{FCEFD8}
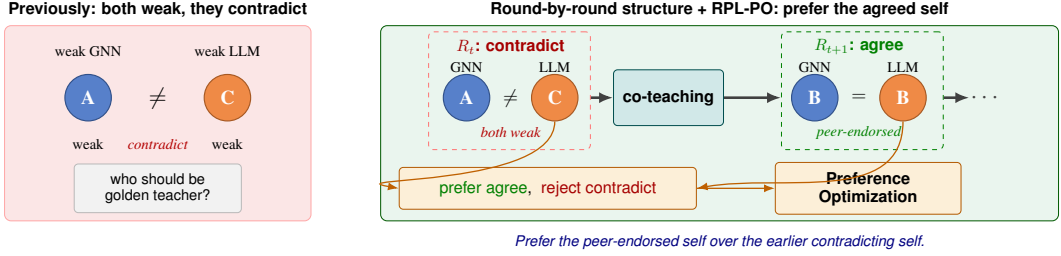
\begin{figure}[t]
\centering
\resizebox{\linewidth}{!}{%
\begin{tikzpicture}[
  >=Latex, thick, font=\sffamily,
  every node/.append style={align=center},
  ptitle/.style={font=\sffamily\bfseries\normalsize},
  panelL/.style={draw=red!35, fill=leftbg, rounded corners=4pt, line width=0.6pt},
  panelR/.style={draw=green!35!black, fill=rightbg, rounded corners=4pt, line width=0.6pt},
  RtFrame/.style={draw=red!55, dashed, rounded corners=3pt, line width=0.6pt},
  RnextFrame/.style={draw=green!50!black, dashed, rounded corners=3pt, line width=0.6pt},
  predGNN/.style={circle, draw=gnnblue!70!black, fill=gnnblue!90, text=white, font=\bfseries\large, minimum size=10mm, line width=0.7pt, inner sep=0pt},
  predLLM/.style={circle, draw=llmorange!70!black, fill=llmorange!90, text=white, font=\bfseries\large, minimum size=10mm, line width=0.7pt, inner sep=0pt},
  midbox/.style={draw=teal!60!black, fill=teal!18, rounded corners=2.5pt, inner sep=2mm, font=\sffamily\bfseries\normalsize, line width=0.6pt, minimum width=22mm, minimum height=15mm},
  prefcell/.style={draw=orange!70!black, fill=prefbg, rounded corners=2.5pt, inner sep=2mm, font=\sffamily\normalsize, line width=0.6pt, minimum height=9mm},
  qbox/.style={draw=gray!55, fill=gray!10, rounded corners=2.5pt, inner sep=2mm, font=\sffamily\small, line width=0.6pt, minimum width=36mm},
  ourarr/.style={->, line width=1.2pt, draw=black!75},
  flowarr/.style={-{Latex[length=3mm,width=2.4mm]}, line width=1.6pt, draw=black!75},
  refarr/.style={->, line width=0.8pt, draw=orange!75!black},
]
\node[ptitle] at (0, 2.45) {Previously: both weak, they contradict};
\node[panelL, minimum width=6.6cm, minimum height=4.2cm] at (0, 0) {};
\node[predGNN] (LA) at (-1.5, 0.55) {A};
\node[predLLM] (LC) at ( 1.5, 0.55) {C};
\node[font=\Large] at (0, 0.55) {$\neq$};
\node[font=\small] at (-1.5, 1.55) {weak GNN};
\node[font=\small] at ( 1.5, 1.55) {weak LLM};
\node[font=\small] at (-1.5, -0.45) {weak};
\node[font=\small] at ( 1.5, -0.45) {weak};
\node[font=\small\itshape, color=red!70!black] at (0, -0.45) {contradict};
\node[qbox] at (0, -1.40) {who should be\\golden teacher?};

\node[ptitle] at (12.1, 2.45) {Round-by-round structure + RPL-PO: prefer the agreed self};
\node[panelR, minimum width=14.6cm, minimum height=4.2cm] at (12.1, 0) {};

\node[font=\sffamily\bfseries\normalsize, color=red!65!black] at (7.6, 1.65) {$R_t$: contradict};
\node[predGNN] (R1G) at (6.65, 0.55) {A};
\node[predLLM] (R1L) at (8.55, 0.55) {C};
\node[font=\large] at (7.6, 0.55) {$\neq$};
\node[font=\small] at (6.65, 1.20) {GNN};
\node[font=\small] at (8.55, 1.20) {LLM};
\node[font=\small\itshape, color=red!70!black] at (7.6, -0.20) {both weak};
\node[RtFrame, fit={(5.95, -0.45) (9.20, 1.85)}] (Rt) {};

\node[midbox, minimum width=18mm, minimum height=12mm] (mid) at (11.0, 0.55) {co-teaching};

\node[font=\sffamily\bfseries\normalsize, color=green!50!black] at (15.10, 1.65) {$R_{t{+}1}$: agree};
\node[predGNN] (R2G) at (14.15, 0.55) {B};
\node[predLLM] (R2L) at (16.05, 0.55) {B};
\node[font=\large] at (15.10, 0.55) {$=$};
\node[font=\small] at (14.15, 1.20) {GNN};
\node[font=\small] at (16.05, 1.20) {LLM};
\node[font=\small\itshape, color=green!50!black] at (15.10, -0.20) {peer-endorsed};
\node[RnextFrame, fit={(13.55, -0.45) (16.80, 1.85)}] (Rnext) {};

\draw[flowarr] (Rt.east |- mid.west) -- (mid.west);
\draw[flowarr] (mid.east) -- (Rnext.west |- mid.east);

\draw[flowarr] (Rnext.east |- mid.east) -- ++(0.55, 0);
\node[font=\Large\bfseries, color=black!75] at (17.85, 0.55) {$\cdots$};

\node[prefcell, minimum width=6.4cm] (prefL) at (8.40, -1.40) {{\color{green!50!black}prefer agree},\ \ {\color{red!65!black}reject contradict}};
\node[prefcell, minimum width=4.0cm, font=\sffamily\bfseries\normalsize] (prefR) at (15.30, -1.40) {Preference\\Optimization};
\draw[refarr] (prefL.east) -- (prefR.west);

\draw[refarr] (R1L.south) to[out=-90, in=170] (prefL.west);
\draw[refarr] (R2L.south) to[out=-90, in=10]  (prefL.east);

\node[font=\sffamily\itshape\small, color=blue!50!black] at (12.1, -2.55)
  {Prefer the peer-endorsed self over the earlier contradicting self.};
\end{tikzpicture}}
\caption{\textbf{Co-teaching without a golden teacher.} A single round between two weak models leaves them contradicting on node $v$, with no way to choose which should serve as the golden teacher (left). After one more round of bidirectional co-teaching, both models evolve, and if they now agree on $B$, the LLM's $R_t$ contradicting answer $C$ and $R_{t+1}$ peer-endorsed answer $B$ together form a preference pair: the earlier self is rejected, the peer-endorsed self is preferred (right). The reward signal comes from the trajectory itself. No golden teacher, no human label, no reward model, no external judge.}
\label{fig:intro_motivation}
\end{figure}

Our answer is \ours{}, a co-teaching framework that does not designate either side as the golden teacher and instead lets the GNN and LLM evolve together. Training proceeds in rounds: in each round, every peer extracts its most confident pseudo-labels under an architecture-specific small-loss criterion (cross-entropy fit for the GNN, minimum token log probability for the LLM) and passes them to the other model, so that both peers grow from weak to strong round by round. To create additional supervision, we further mine a preference signal from this trajectory: whenever a node transitions from cross-model contradiction at round $t$ to cross-model agreement at round $t+1$, the LLM's two answers on the same node, the earlier contradicting one and the later peer-endorsed one, form a natural preference pair, which we feed to direct preference optimization (DPO)~\citep{rafailov2023direct}. We call this Round-based Pseudo-Label Preference Optimization (RPL-PO). The reward signal comes from the trajectory itself: no golden teacher, no human label, no reward model, no external judge.

\paragraph{Contributions.} (1)~\textbf{We abandon the golden-teacher assumption.} \ours{} is the first LLM-GNN method in which neither model is designated as authoritative, with both updating every round and supervising each other through a small-loss criterion. (2)~\textbf{RPL-PO: a self-supervised preference-pair generator.} A node that transitions from cross-model contradiction at round $t$ to cross-model agreement at round $t+1$ yields a DPO preference pair from the LLM's two answers on the same input. RPL-PO requires no human labels, no reward models, and no external judges, and is structurally inaccessible to single-round or frozen-teacher pipelines. (3)~\textbf{State-of-the-art on six benchmarks.} \ours{} outperforms GNN-as-Judge by up to $7.86$ percentage points on Cora and $7.73$ percentage points on ogbn-arxiv under 3-shot supervision, with the same lead carrying over to 5-shot and to zero-shot cross-dataset transfer. The error-structure analysis in \S\ref{sec:error_structure} shows that abandoning the golden-teacher assumption substantially improves the LLM's graph learning capability on challenging samples.

\section{Related Work}
\label{sec:related}

\paragraph{LLM-GNN methods for graph learning.} Combining LLMs and GNNs for TAGs has been extensively explored~\citep{chen2023label,tang2024graphgpt,ye2024language,he2023harnessing,guo2023gpt4graph,huang2023can,wu2025llms,dai2024large,wen2023augmenting,yu2025leveraging}. \emph{LLM-as-Enhancers}~\citep{he2023harnessing,zhao2022learning,li2023grenade,yang2021graphformers} freeze LLM-derived features or explanations as enriched node input to a downstream GNN. \emph{LLM-as-Predictors}~\citep{tang2024graphgpt,chen2023label,ye2024language,chen2024llaga,liu2023one,hu2024let,wang2024llms,wang2024instructgraph} frame node classification as text generation, typically with structural prompts or graph tokens. \emph{GNN-as-Judge}~\citep{xu2026gnn} reverses the direction: a once-trained GNN's verdicts filter pseudo-labels for fine-tuning the LLM, with a theoretical lower bound on agreement-set accuracy under conditional independence; \citet{sheng2025llms} similarly treats LLM annotations as noisy oracles for graph active learning. In every case, one model is fixed as the golden teacher and supervision flows in one direction. \ours{} instead designates no golden teacher: both models update every round and judge each other across multiple rounds.

\paragraph{Co-teaching, noisy labels, and pseudo-label selection.} Co-teaching~\citep{han2018co} trains two networks simultaneously, each selecting small-loss samples for its peer. Co-Teaching+~\citep{yu2019does} adds disagreement filtering, DivideMix~\citep{li2020dividemix} introduces mixture-model selection, and earlier co-training~\citep{qiao2018deep,ma2017self,kumar2010self} variants pair networks of the same architecture. The broader noisy-label literature~\citep{natarajan2013noisy,gui2021towards,chen2019understanding,cheng2020learning,luo2024robustft} likewise treats noise as homogeneous across views, and recent work warns that LLMs trained on their own outputs can degrade over time~\citep{shumailov2023curse}. Closely related is pseudo-labeling~\citep{lee2013pseudo}, which augments small labeled sets with model-generated labels, with mining of both easy and hard samples shown to be crucial~\citep{mukherjee2020uncertainty,rizve2021defense}. On graphs, prior work explores multi-stage self-training~\citep{sun2020multi}, label-propagation hybrids~\citep{wang2020unifying,li2018deeper}, confidence-aware filtering~\citep{liu2022confidence}, and active labeling~\citep{cai2017active,zhang2021rim}, with single-round GNN-LLM agreement filters~\citep{xu2026gnn} the closest to our setup. All of this prior work pairs homogeneous networks of the same architecture and uses a single-round selection. We are the first to co-teach across \emph{heterogeneous} architectures (GNN + LLM) iteratively, whose complementary inductive biases (structural vs.\ semantic) provide stronger error independence than random-initialization diversity.

\paragraph{Preference optimization.} LLM alignment from feedback originates in RLHF~\citep{christiano2017deep,stiennon2020learning,ouyang2022training}, with DPO~\citep{rafailov2023direct} and its variants~\citep{azar2024general,ethayarajh2024model,meng2024simpo,zhao2023group,amini2024direct,hong2024orpo} replacing the reward model with pairwise preferences. On graphs, GNN-as-Judge~\citep{xu2026gnn} and InstructGraph~\citep{wang2024instructgraph} apply preference tuning to within-round GNN-LLM disagreements; RPL-PO instead exploits the \emph{temporal} structure of co-teaching, contrasting the same LLM's predictions across consecutive rounds.

\paragraph{Our novelty.} \ours{} contributes on two fronts. (i) \emph{A new LLM-GNN framework without an explicitly designated golden teacher}, operationalized via heterogeneous co-teaching in which the GNN and LLM iteratively pseudo-label each other under a small-loss criterion and both update every round. (ii) \emph{A novel preference optimization signal mined from the learning trajectory}, which constructs preference pairs from cross-round agreement transitions and requires no golden teacher, no human label, no reward model, and no external judge, fully releasing the supervision signal latent in sparse-label graph learning.

\section{Preliminaries}
\label{sec:prelim}

We consider few-shot semi-supervised node classification on text-attributed graphs (TAGs) defined as $\mathcal{G} = (\mathcal{V}, \mathcal{E}, \mathbf{T})$, where $\mathcal{V} = \{v_1, v_2, \ldots, v_N\}$ is the node set, $\mathcal{E} \subseteq \mathcal{V}\times\mathcal{V}$ is the edge set, and $\mathbf{T} = \{t_v\}_{v\in\mathcal{V}}$ are the per-node text attributes (e.g.\ paper title and abstract in citation networks). Each node $v$ has a label $y_v \in \{1,\ldots,C\}$. Given a small labeled set $\mathcal{V}_{\text{train}}$ in which each class has exactly $k$ labeled nodes, along with a validation set $\mathcal{V}_{\text{val}}$, the goal is to predict labels for the test nodes $\mathcal{V}_{\text{test}} = \mathcal{V} \setminus (\mathcal{V}_{\text{train}}\cup\mathcal{V}_{\text{val}})$. Few-shot \emph{semi-supervised} learning trains the models on both (i) the small labeled set $\mathcal{V}_{\text{train}}$ together with its ground-truth labels and (ii) the unlabeled nodes $\mathcal{V}_{\text{U}} \subseteq \mathcal{V}\setminus\mathcal{V}_{\text{train}}$ and their text contents, e.g., by predicting on $\mathcal{V}_{\text{U}}$ and using the most confident predictions as pseudo-labels.

\paragraph{LLM predictor.} A large language model $f_{\text{LLM}}$ predicts node labels by framing classification as text generation. Given node $v$, we build a prompt $\mathcal{P}(v)$ containing $v$'s text $t_v$, the texts of $v$'s neighbors, and the candidate class names. The LLM emits a class name, giving prediction $\hat y_v^L = f_{\text{LLM}}(\mathcal{P}(v))$.

\paragraph{GNN predictor.} A graph neural network $f_{\text{GNN}}: \mathcal{V}\to\mathbb{R}^C$ produces class logits by aggregating information from a node's local neighborhood via message passing. Layer $\ell$ computes
\begin{equation}
\mathbf{h}_v^{(\ell)} = \text{UPDATE}^{(\ell)}\!\left(\mathbf{h}_v^{(\ell-1)},\; \text{AGG}^{(\ell)}\!\left(\{\mathbf{h}_u^{(\ell-1)}: u\in\mathcal{N}(v)\}\right)\right),
\label{eq:gnn}
\end{equation}
with $\mathbf{h}_v^{(0)} = \mathbf{x}_v$ (a numerical embedding of $t_v$) and final prediction $\hat y_v^G = \arg\max f_{\text{GNN}}(v)$.

\section{LLM-GNN Co-teaching}
\label{sec:method}

\textbf{Why no golden teacher?} Existing LLM-GNN methods all designate one model as a fixed teacher whose outputs are treated as ground truth. Under sparse supervision, this assumption breaks down: with few labels per class, neither model is reliable, so freezing either side as the teacher transfers its blind spots into the student wholesale and the unidirectional supervision leaves no path to revise mistakes. Effective few-shot LLM-GNN learning therefore needs a framework where neither model is fixed as teacher and both can correct each other through their joint dynamics.

\textbf{Difference from classical co-teaching.} Classical Co-Teaching~\citep{han2018co} pairs two networks of the same architecture and relies on random-init diversity for the two peers to make different mistakes; this diversity is fragile and shrinks as the two peers converge. \ours{} instead pairs a GNN and an LLM, whose inductive biases are disjoint by construction: the LLM's text-only view fails on semantically ambiguous descriptions, while the GNN's structural view fails on low-degree, sparsely connected nodes. Each peer is reliably strong where the other is weak, so confident pseudo-labels from one carry information the other could not have produced alone, providing a far more durable basis for mutual correction than random initialization. \S\ref{sec:error_structure} confirms this complementarity empirically through per-degree error rates, and Appendix~\ref{app:venn_per_dataset} shows it through per-dataset Venn diagrams of LLM/GNN correctness.

\ours{} runs for $T$ rounds. Figure~\ref{fig:method_flow} visualises one round, organised in three stages.
\begin{enumerate}[nosep,leftmargin=1.5em]
\item \textbf{Select and exchange confident pseudo-labels} (\S\ref{sec:coteaching}). Both models predict on the same unlabeled batch, each picks its most confident subset, and the two subsets are swapped.
\item \textbf{Update the LLM and GNN} (\S\ref{sec:llm_opt}). The LLM is fine-tuned on the GNN-selected pseudo-labels and the GNN is trained on the LLM-selected pseudo-labels, both starting from the previous round's checkpoint.
\item \textbf{Reinforce pseudo-label quality with RPL-PO} (\S\ref{sec:rplpo}). Every two rounds, nodes whose LLM prediction flipped from disagreeing-with-GNN to agreeing-with-GNN form a temporal preference pair. The LLM is then updated by DPO on these pairs, with no human annotation or external reward model.
\end{enumerate}

\definecolor{flowgnn}{HTML}{4472C4}
\definecolor{flowllm}{HTML}{ED7D31}
\begin{figure}[t]
\centering
\resizebox{\linewidth}{!}{%
\begin{tikzpicture}[
  >=Latex, thick, font=\small,
  every node/.append style={align=center},
  io/.style={draw, rounded corners=2.5pt, minimum width=22mm, minimum height=10mm, fill=gray!12},
  gnn/.style={draw=flowgnn!75!black, fill=flowgnn!18, rounded corners=2.5pt, minimum width=33mm, minimum height=11mm},
  llm/.style={draw=flowllm!75!black, fill=flowllm!28, rounded corners=2.5pt, minimum width=33mm, minimum height=11mm},
  rplbadge/.style={draw=orange!60!black, fill=orange!10, rounded corners=2.5pt, minimum width=42mm, minimum height=10mm, dashed},
  arr/.style={->, thick, draw=black!70},
  cross/.style={->, very thick, draw=red!70!black},
]
\node[io] (B) at (0,0) {Sample batch\\$\mathcal{B}_t \subset \mathcal{V}_\text{U}$};
\node[llm, right=10mm of B, yshift=11mm] (L2) {{\bfseries LLM} predicts\\$\hat y_v^L$ on $\mathcal{B}_t$};
\node[gnn, right=10mm of B, yshift=-11mm] (G2) {{\bfseries GNN} predicts\\$\hat y_v^G$ on $\mathcal{B}_t$};
\node[llm, right=8mm of L2] (L3) {top-$R(t)$ confident\\$\mathcal{S}_t^{L\to G}$};
\node[gnn, right=8mm of G2] (G3) {top-$R(t)$ confident\\$\mathcal{S}_t^{G\to L}$};
\node[llm, right=14mm of L3] (L4) {{\bfseries LLM} SFT on\\$\mathcal{V}_{\text{train}} \cup \mathcal{S}_t^{G\to L}$};
\node[gnn, right=14mm of G3] (G4) {{\bfseries GNN} train on\\$\mathcal{V}_{\text{train}} \cup \mathcal{S}_t^{L\to G}$};
\node[rplbadge, right=10mm of L4] (rplbadge) {{\bfseries RPL-PO}\\(every 2 rounds, see below)};

\draw[arr] (B) -- (L2);
\draw[arr] (B) -- (G2);
\draw[arr] (L2) -- (L3);
\draw[arr] (G2) -- (G3);
\draw[cross] (L3.east) -- (G4.west);
\draw[cross] (G3.east) -- (L4.west);
\node[font=\normalsize\itshape, color=red!70!black, fill=white, inner sep=1pt]
  at ($(L3.east)!0.5!(G4.west)$) {exchange labels};
\draw[arr, dashed, draw=orange!60!black] (L4.east) -- (rplbadge.west);

\begin{scope}[on background layer]
  \node[draw=black, dashed, line width=1pt, rounded corners=4pt, fit=(B)(L2)(G2)(L3)(G3), inner sep=3mm] (blockA) {};
  \node[draw=black, dashed, line width=1pt, rounded corners=4pt, fit=(L4)(G4), inner sep=3mm] (blockB) {};
  \node[draw=black, dashed, line width=1pt, rounded corners=4pt, fit=(rplbadge), inner sep=3mm] (blockC) {};
\end{scope}
\node[font=\normalsize\bfseries, below=0.5mm of blockA.south] {Select \& exchange confident pseudo-labels};
\node[font=\normalsize\bfseries, below=0.5mm of blockB.south] {Update LLM and GNN};
\node[font=\normalsize\bfseries, below=0.5mm of blockC.south] {Reinforce pseudo-label quality};
\end{tikzpicture}}
\caption{Round-$t$ co-teaching flow. The \textcolor{flowgnn!75!black}{\bfseries GNN} and \textcolor{flowllm!75!black}{\bfseries LLM} predict on the same unlabeled batch $\mathcal{B}_t$ and score by a model-specific self-confidence signal. The top-$R(t)$ confident pairs are \textcolor{red!70!black}{cross-exchanged}: $\mathcal{S}_t^{G\to L}$ supervises the LLM in \S\ref{sec:llm_opt}, and $\mathcal{S}_t^{L\to G}$ supervises the GNN in \S\ref{sec:gnn_opt}. Every two rounds the LLM additionally undergoes \textcolor{orange!70!black}{RPL-PO} (zoomed in Fig.~\ref{fig:rplpo_zoom}, \S\ref{sec:rplpo}).}
\label{fig:method_flow}
\end{figure}
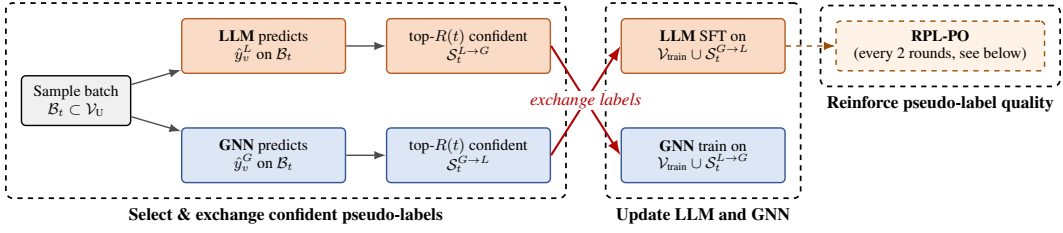

\subsection{Selecting and Exchanging Confident Pseudo-Labels}
\label{sec:coteaching}

\textbf{Intuition.} Two weak models can still teach each other if they are honest about what they do not know. We therefore let each model speak only on the nodes where it is confident, and we send those confident answers to the peer as supervision; the peer treats them as ground truth on \emph{its} blind spots. Concretely, each model passes to its peer only the pseudo-labels it is most certain about. Certainty is measured by how strongly the model's own scoring agrees with the answer it just produced, following the small-loss principle of Co-Teaching~\citep{han2018co}. An LLM that hesitates on some token of its generated answer, or a GNN that poorly fits its own predicted class, is more likely to be wrong than a model that scores its own answer with high confidence.

For each node $v \in \mathcal{B}_t$ we compute a self-confidence signal,
\[
\ell_v^G \;=\; \ell_\text{CE}\!\bigl(f_{\text{GNN}}(v),\,\hat y_v^G\bigr) \quad\text{(GNN)}, \qquad
d_v \;=\; \min_{j}\,\log p_{\text{LLM}}(w_j\mid w_{<j}) \quad\text{(LLM)},
\]
where low $\ell_v^G$ means the GNN strongly fits the label it just produced and high $d_v$ means no token along the LLM's produced answer was uncertain. We keep the most confident top-$R(t)$ fraction. The selection ratio $R(t) \in (0, 1]$ is annealed linearly from $R_{\min}$ at $t{=}1$ to $R_{\max}$ at $t{=}T$,
\begin{equation}
R(t) \;=\; R_{\min} + (R_{\max} - R_{\min})\cdot \frac{t-1}{T-1},
\label{eq:rt}
\end{equation}
so early rounds (when both models are still weak) exchange few but reliable pseudo-labels, while later rounds exchange more. The two paired sets $\mathcal{S}_t^{G\to L} = \{(v,\,\hat y_v^G)\}$ and $\mathcal{S}_t^{L\to G} = \{(v,\,\hat y_v^L)\}$ resulting from this exchange feed the model updates that follow.

\subsection{Updating the LLM and GNN}
\label{sec:llm_opt}
\label{sec:gnn_opt}

\textbf{Intuition.} The exchanged pseudo-labels are the peer's best guesses on the receiver's weak spots, so each model now has access to supervision it could not have produced alone. We update both models on the same training data structure, anchors plus peer-supplied pseudo-labels, but with model-specific losses. Both updates start from the previous round's checkpoint, so each model carries forward the gains from earlier rounds.

\paragraph{LLM SFT.} At round $t$, the LLM $f_{\text{LLM}}$ is fine-tuned on the labeled anchors $\mathcal{V}_{\text{train}}$ together with the GNN-selected pseudo-label set $\mathcal{S}_t^{G\to L}$, starting from the previous round's adapter. Let $\hat y$ denote the supervision target: the ground-truth label $y_v$ when $v\in\mathcal{V}_{\text{train}}$, and the GNN's prediction $\hat y_v^G$ when $v\in\mathcal{S}_t^{G\to L}$. Let $\hat y_{<i}$ denote its first $i-1$ tokens. We minimise the standard token-level cross-entropy
\begin{equation}
\mathcal{L}_{\text{LLM}}^{(t)} \;=\; -\!\!\sum_{(v,\hat y)\,\in\,\mathcal{V}_{\text{train}}\,\cup\,\mathcal{S}_t^{G\to L}}\;\sum_{i=1}^{|\hat y|} \log p_{\text{LLM}}\!\big(\hat y_i \mid \mathcal{P}(v),\,\hat y_{<i}\big),
\label{eq:llm_loss}
\end{equation}
with $p_{\text{LLM}}$ the LLM's next-token distribution and $\mathcal{P}(v)$ the prompt defined in \S\ref{sec:prelim}.

\paragraph{GNN training.} The GNN is trained on the few ground-truth anchors and the larger LLM-selected pseudo-label set. Naively concatenating the two would let the larger pseudo-label set drown out the anchor signal, so we average each loss within its own set and combine them with a round-dependent weight. This gives a convex combination of an anchor loss and a pseudo-label loss:
\begin{equation}
\mathcal{L}_{\text{GNN}}^{(t)} \;=\; (1-\alpha_t)\cdot\underbrace{\tfrac{1}{|\mathcal{V}_{\text{train}}|}\!\!\sum_{(v,y)\in\mathcal{V}_{\text{train}}}\!\!\ell_\text{CE}(f_{\text{GNN}}(v),y)}_{\text{anchor loss}}
\;+\; \alpha_t\cdot\underbrace{\tfrac{1}{|\mathcal{S}_t^{L\to G}|}\!\!\sum_{(v,\hat y_v^L)\in\mathcal{S}_t^{L\to G}}\!\!\ell_\text{CE}(f_{\text{GNN}}(v),\hat y_v^L)}_{\text{pseudo-label loss}}.
\label{eq:gnn_loss}
\end{equation}
The mixing weight $\alpha_t$ is annealed linearly from $\alpha_0$ at round 1 to $\alpha_{\max}$ at round $T$. Early rounds emphasise the anchor signal. Later rounds, when the LLM produces cleaner pseudo-labels, weight those more heavily, while the within-set averaging prevents the larger pseudo-label set from washing out the anchors.

\subsection{Reinforcing Pseudo-Label Quality with RPL-PO}
\label{sec:rplpo}

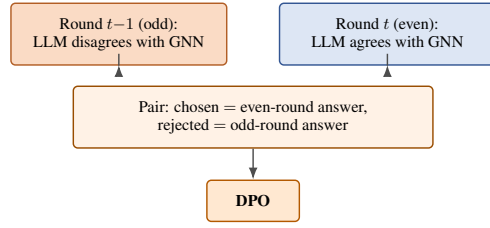
\begin{wrapfigure}{r}{0.46\textwidth}
\vspace{-0.6\baselineskip}
\centering
\resizebox{\linewidth}{!}{%
\begin{tikzpicture}[
  >=Latex, thick, font=\small,
  every node/.append style={align=center},
  rd/.style={draw, rounded corners=2.5pt, minimum width=42mm, minimum height=12mm, fill=gray!8},
  oddbox/.style={rd, draw=flowllm!75!black, fill=flowllm!28},
  evenbox/.style={rd, draw=flowgnn!75!black, fill=flowgnn!18},
  pairbox/.style={draw=orange!60!black, fill=orange!10, rounded corners=2.5pt, minimum width=70mm, minimum height=12mm},
  dpobox/.style={draw=orange!70!black, fill=orange!18, rounded corners=2.5pt, minimum width=18mm, minimum height=8mm},
  arr/.style={->, thick, draw=black!70},
]
\node[oddbox] (odd) {Round $t{-}1$ (odd):\\LLM disagrees with GNN};
\node[evenbox, right=10mm of odd] (even) {Round $t$ (even):\\LLM agrees with GNN};
\node[pairbox, below=10mm of $(odd)!0.5!(even)$] (pair) {Pair: chosen $=$ even-round answer,\\rejected $=$ odd-round answer};
\node[dpobox, below=6mm of pair] (dpo) {{\bfseries DPO}};
\draw[arr] (odd.south) |- ([yshift=4mm]pair.north -| odd.south);
\draw[arr] (even.south) |- ([yshift=4mm]pair.north -| even.south);
\draw[arr] (pair) -- (dpo);
\end{tikzpicture}}
\caption{RPL-PO zoom-in: every two rounds, the LLM's even-round answer (peer-confirmed) is preferred.}
\label{fig:rplpo_zoom}
\vspace{-0.6\baselineskip}
\end{wrapfigure}
We incorporate RPL-PO to mine an additional supervisory signal from the cross-round trajectory itself: the same LLM's two answers on the same node, before and after one round of teaching, form a free preference pair without any label or external judge.

\paragraph{Intuition.} Round-by-round co-teaching gives us something single-round pipelines do not have: \emph{the same LLM's two answers on the same node, before and after one round of teaching}. When the second answer is endorsed by the now-stronger GNN while the first was not, the LLM has visibly self-corrected on that node, and we want training to make this correction stick. We harvest this signal as \emph{Round-based Pseudo-Label Preference Optimization} (RPL-PO), illustrated in Fig.~\ref{fig:rplpo_zoom}: the later, peer-endorsed answer is preferred over the earlier, peer-rejected answer, and the LLM is updated by DPO~\citep{rafailov2023direct} on these pairs. As argued in \S\ref{sec:intro}, this lets us mine supervision from the trajectory itself, without designating either model as the golden teacher.

Consecutive rounds $(2k{-}1, 2k)$ are seeded to draw the \emph{same} batch, so for every node $v$ we have an odd-round LLM prediction $\hat y_v^{L,\text{odd}}$ and an even-round prediction $\hat y_v^{L,\text{even}}$ on the same input. We keep $v$ in the preference set when both conditions hold:
\begin{equation}
\underbrace{\hat y_v^{L,\text{odd}} \neq \hat y_v^{G,\text{odd}}}_{\text{(i) odd round disagrees}}
\quad \text{and} \quad
\underbrace{\hat y_v^{L,\text{even}} = \hat y_v^{G,\text{even}}\ \ \text{and}\ \ \hat y_v^{L,\text{even}} \neq \hat y_v^{L,\text{odd}}}_{\text{(ii) LLM changes to agree}}.
\label{eq:rplpo_conds}
\end{equation}
The second condition ensures the pair represents genuine self-correction by the LLM rather than the GNN drifting onto an unchanged LLM. The LLM updated its answer, and the now-stronger GNN endorses the new one. We set
\begin{equation}
\text{chosen} \;=\; \hat y_v^{L,\text{even}}, \qquad \text{rejected} \;=\; \hat y_v^{L,\text{odd}},
\label{eq:rplpo_pair}
\end{equation}
and apply standard DPO~\citep{rafailov2023direct} with the SFT checkpoint as the reference. The contrast is across \emph{time} (same model, two training stages) and grounded by cross-model consensus, providing supervision that requires no human annotation and no external reward model.

\section{Experiments}
\label{sec:experiments}

\subsection{Experimental Setup}

\paragraph{Datasets.} We evaluate on six text-attributed graphs spanning citation networks (Cora~\citep{yang2016revisiting}, Citeseer~\citep{sen2008collective}, PubMed~\citep{sen2008collective}, ogbn-arxiv~\citep{hu2020open}), Wikipedia hyperlinks (WikiCS~\citep{mernyei2020wiki}), and an Amazon product co-purchase subset (ogbn-products~\citep{hu2020open}). The first five overlap with the benchmarks of \citet{xu2026gnn}. We additionally include WikiCS, on which their numbers are not available, and we re-run the published implementations of GNN-as-Judge and the recent LLM-as-Predictor baselines (LLM-GNN, LLaGA, GraphGPT) to fill the WikiCS column. Full per-dataset statistics (\#nodes, \#edges, \#features, \#classes) and the exact $k$-shot training / validation / test splits we use are listed in Appendix~\ref{app:datasets}.

\paragraph{Baselines.} We compare against methods from three categories. (1) \emph{Classical GNN models}: GCN~\citep{kipf2016semi}, GAT~\citep{velickovic2018graph}, and GraphSAGE~\citep{hamilton2017inductive}. (2) \emph{LLM-as-Predictors}: Zero-shot, Graph-CoT~\citep{wei2022chain}, and neighbor-augmented prompting. (3) \emph{LLM-Graph methods}: GLEM~\citep{zhao2022learning}, TAPE~\citep{he2023harnessing}, LLM-GNN~\citep{chen2023label}, LLaGA~\citep{chen2024llaga}, GraphGPT~\citep{tang2024graphgpt}, and GNN-as-Judge~\citep{xu2026gnn}.

\paragraph{Implementation details.} Full implementation details are deferred to Appendix~\ref{app:hyperparameters}. For GLEM \cite{he2023harnessing}, TAPE \cite{he2023harnessing}, LLM-GNN \cite{chen2023label}, LLaGA \cite{chen2024llaga}, GraphGPT \cite{tang2024graphgpt}, and GNN-as-Judge \cite{xu2026gnn}, we verified that the results from \cite{xu2026gnn} are reproducible under matched splits, and report their numbers on Cora, Citeseer, PubMed, ogbn-arxiv, and ogbn-products. WikiCS is not covered by \cite{xu2026gnn} and is re-implemented by us. Other remaining baselines are implemented by us.

\subsection{Few-Shot Semi-Supervised Node Classification}

\begin{table}[t]
\caption{Node classification accuracy (\%) on six benchmarks under 3/5/10-shot settings. \textbf{Bold}: best.}
\label{tab:main}
\centering
\footnotesize
\setlength{\tabcolsep}{3pt}
\renewcommand{\arraystretch}{0.95}
\begin{tabular}{@{}l cccccc@{}}
\toprule
\textbf{Method} & \textbf{Cora} & \textbf{Citeseer} & \textbf{PubMed} & \textbf{WikiCS} & \textbf{ogbn-arxiv} & \textbf{ogbn-products} \\
\midrule
\multicolumn{7}{c}{\textit{3-shot (3 labels per class)}} \\
\midrule
GCN  & 70.72{\scriptsize$\pm$3.06} & 55.14{\scriptsize$\pm$3.82} & 67.72{\scriptsize$\pm$5.19} & 60.33{\scriptsize$\pm$4.59} & 39.83{\scriptsize$\pm$1.48} & 60.14{\scriptsize$\pm$0.41} \\
GAT  & 70.74{\scriptsize$\pm$2.12} & 60.46{\scriptsize$\pm$4.46} & 66.50{\scriptsize$\pm$5.09} & 59.65{\scriptsize$\pm$5.21} & 32.43{\scriptsize$\pm$1.56} & 57.14{\scriptsize$\pm$1.23} \\
SAGE & 68.89{\scriptsize$\pm$2.87} & 56.74{\scriptsize$\pm$3.22} & 66.10{\scriptsize$\pm$5.56} & 58.55{\scriptsize$\pm$4.41} & 34.69{\scriptsize$\pm$2.47} & 58.85{\scriptsize$\pm$0.88} \\
\midrule
Zero-Shot   & 66.15{\scriptsize$\pm$0.36} & 58.56{\scriptsize$\pm$0.84} & 74.78{\scriptsize$\pm$0.65} & 67.65{\scriptsize$\pm$0.75} & 50.35{\scriptsize$\pm$1.66} & 75.88{\scriptsize$\pm$1.05} \\
Graph-CoT   & 63.23{\scriptsize$\pm$0.88} & 48.56{\scriptsize$\pm$2.77} & 86.90{\scriptsize$\pm$2.60} & 74.50{\scriptsize$\pm$1.15} & 49.45{\scriptsize$\pm$1.05} & 74.56{\scriptsize$\pm$1.95} \\
w. Neighbor & 68.86{\scriptsize$\pm$1.78} & 54.05{\scriptsize$\pm$1.55} & 75.26{\scriptsize$\pm$2.56} & 72.05{\scriptsize$\pm$2.54} & 49.50{\scriptsize$\pm$1.86} & 76.95{\scriptsize$\pm$1.55} \\
\midrule
GLEM     & 67.81{\scriptsize$\pm$0.92} & 53.09{\scriptsize$\pm$1.39} & 63.85{\scriptsize$\pm$1.02} & 71.56{\scriptsize$\pm$1.56} & 36.37{\scriptsize$\pm$1.96} & 52.46{\scriptsize$\pm$1.93} \\
TAPE     & 73.71{\scriptsize$\pm$1.86} & 64.96{\scriptsize$\pm$0.36} & 71.33{\scriptsize$\pm$0.87} & 73.56{\scriptsize$\pm$2.55} & 48.25{\scriptsize$\pm$0.77} & 69.64{\scriptsize$\pm$1.15} \\
LLM-GNN  & 73.85{\scriptsize$\pm$0.74} & 61.67{\scriptsize$\pm$0.58} & 66.01{\scriptsize$\pm$0.63} & 72.15{\scriptsize$\pm$2.56} & 42.36{\scriptsize$\pm$1.72} & 55.46{\scriptsize$\pm$0.37} \\
LLaGA    & 54.79{\scriptsize$\pm$0.91} & 32.93{\scriptsize$\pm$1.24} & 43.96{\scriptsize$\pm$1.74} & 71.56{\scriptsize$\pm$0.56} & 29.73{\scriptsize$\pm$2.82} & 30.67{\scriptsize$\pm$1.54} \\
GraphGPT & 57.77{\scriptsize$\pm$1.62} & 52.34{\scriptsize$\pm$1.06} & 57.51{\scriptsize$\pm$1.89} & 72.26{\scriptsize$\pm$1.68} & 31.26{\scriptsize$\pm$2.73} & 40.83{\scriptsize$\pm$2.97} \\
\midrule
GNN-as-Judge & 77.89{\scriptsize$\pm$1.28} & 73.59{\scriptsize$\pm$0.64} & 87.12{\scriptsize$\pm$0.89} & 67.50{\scriptsize$\pm$1.02} & 62.21{\scriptsize$\pm$1.45} & 81.02{\scriptsize$\pm$1.23} \\
\textbf{LG Co-Teaching} & \best{85.75}{\scriptsize$\pm$0.88} & \best{77.12}{\scriptsize$\pm$0.97} & \best{91.32}{\scriptsize$\pm$1.56} & \best{74.80}{\scriptsize$\pm$0.95} & \best{69.94}{\scriptsize$\pm$0.84} & \best{82.82}{\scriptsize$\pm$1.02} \\
\midrule
\multicolumn{7}{c}{\textit{5-shot (5 labels per class)}} \\
\midrule
GCN  & 75.99{\scriptsize$\pm$1.95} & 62.30{\scriptsize$\pm$3.50} & 74.05{\scriptsize$\pm$1.78} & 66.14{\scriptsize$\pm$2.76} & 44.96{\scriptsize$\pm$1.33} & 68.33{\scriptsize$\pm$0.22} \\
GAT  & 74.70{\scriptsize$\pm$2.35} & 65.56{\scriptsize$\pm$2.12} & 72.32{\scriptsize$\pm$1.52} & 65.32{\scriptsize$\pm$1.58} & 39.27{\scriptsize$\pm$1.19} & 66.80{\scriptsize$\pm$0.45} \\
SAGE & 74.05{\scriptsize$\pm$2.78} & 64.11{\scriptsize$\pm$2.73} & 71.85{\scriptsize$\pm$2.13} & 64.39{\scriptsize$\pm$2.89} & 39.75{\scriptsize$\pm$3.23} & 67.55{\scriptsize$\pm$0.67} \\
\midrule
Zero-Shot   & 66.15{\scriptsize$\pm$0.36} & 58.56{\scriptsize$\pm$0.84} & 74.78{\scriptsize$\pm$0.65} & 67.65{\scriptsize$\pm$0.75} & 50.35{\scriptsize$\pm$1.66} & 75.88{\scriptsize$\pm$1.05} \\
Graph-CoT   & 63.23{\scriptsize$\pm$0.88} & 48.56{\scriptsize$\pm$2.77} & 86.90{\scriptsize$\pm$2.60} & 74.50{\scriptsize$\pm$1.15} & 49.45{\scriptsize$\pm$1.05} & 74.56{\scriptsize$\pm$1.95} \\
w. Neighbor & 68.86{\scriptsize$\pm$1.78} & 54.05{\scriptsize$\pm$1.55} & 75.26{\scriptsize$\pm$2.56} & 72.05{\scriptsize$\pm$2.54} & 49.50{\scriptsize$\pm$1.86} & 76.95{\scriptsize$\pm$1.55} \\
\midrule
GLEM     & 74.69{\scriptsize$\pm$0.46} & 61.71{\scriptsize$\pm$0.58} & 73.29{\scriptsize$\pm$1.44} & 75.56{\scriptsize$\pm$1.84} & 39.19{\scriptsize$\pm$1.57} & 56.87{\scriptsize$\pm$1.70} \\
TAPE     & 74.28{\scriptsize$\pm$0.81} & 67.73{\scriptsize$\pm$0.48} & 75.02{\scriptsize$\pm$0.83} & 76.15{\scriptsize$\pm$2.11} & 55.22{\scriptsize$\pm$0.48} & 77.44{\scriptsize$\pm$1.32} \\
LLM-GNN  & 75.61{\scriptsize$\pm$1.04} & 62.37{\scriptsize$\pm$1.35} & 74.33{\scriptsize$\pm$0.95} & 76.32{\scriptsize$\pm$1.23} & 45.74{\scriptsize$\pm$1.66} & 64.01{\scriptsize$\pm$0.39} \\
LLaGA    & 62.88{\scriptsize$\pm$2.19} & 43.71{\scriptsize$\pm$4.36} & 58.63{\scriptsize$\pm$1.05} & 74.11{\scriptsize$\pm$2.31} & 33.74{\scriptsize$\pm$2.45} & 37.29{\scriptsize$\pm$3.12} \\
GraphGPT & 60.17{\scriptsize$\pm$1.44} & 51.83{\scriptsize$\pm$2.24} & 57.39{\scriptsize$\pm$3.67} & 75.16{\scriptsize$\pm$1.22} & 36.25{\scriptsize$\pm$1.87} & 44.78{\scriptsize$\pm$2.34} \\
\midrule
GNN-as-Judge & 79.54{\scriptsize$\pm$0.39} & 74.39{\scriptsize$\pm$1.63} & 87.49{\scriptsize$\pm$1.23} & 78.10{\scriptsize$\pm$1.23} & 66.76{\scriptsize$\pm$0.83} & 81.93{\scriptsize$\pm$2.21} \\
\textbf{LG Co-Teaching} & \best{86.10}{\scriptsize$\pm$1.32} & \best{77.32}{\scriptsize$\pm$0.98} & \best{91.85}{\scriptsize$\pm$1.12} & \best{78.33}{\scriptsize$\pm$0.68} & \best{70.05}{\scriptsize$\pm$1.22} & \best{83.45}{\scriptsize$\pm$2.10} \\
\midrule
\multicolumn{7}{c}{\textit{10-shot (10 labels per class)}} \\
\midrule
GCN  & 79.40{\scriptsize$\pm$0.98} & 67.55{\scriptsize$\pm$2.57} & 76.24{\scriptsize$\pm$2.88} & 71.66{\scriptsize$\pm$1.27} & 50.29{\scriptsize$\pm$1.50} & 70.24{\scriptsize$\pm$0.24} \\
GAT  & 77.41{\scriptsize$\pm$1.51} & 67.84{\scriptsize$\pm$1.60} & 74.87{\scriptsize$\pm$3.04} & 70.43{\scriptsize$\pm$0.74} & 45.49{\scriptsize$\pm$0.87} & 67.97{\scriptsize$\pm$0.65} \\
SAGE & 78.24{\scriptsize$\pm$1.11} & 67.73{\scriptsize$\pm$2.30} & 74.88{\scriptsize$\pm$1.96} & 71.26{\scriptsize$\pm$0.93} & 45.24{\scriptsize$\pm$1.33} & 68.99{\scriptsize$\pm$0.41} \\
\midrule
Zero-Shot   & 66.15{\scriptsize$\pm$0.36} & 58.56{\scriptsize$\pm$0.84} & 74.78{\scriptsize$\pm$0.65} & 67.65{\scriptsize$\pm$0.75} & 50.35{\scriptsize$\pm$1.66} & 75.88{\scriptsize$\pm$1.05} \\
Graph-CoT   & 63.23{\scriptsize$\pm$0.88} & 48.56{\scriptsize$\pm$2.77} & 86.90{\scriptsize$\pm$2.60} & 74.50{\scriptsize$\pm$1.15} & 49.45{\scriptsize$\pm$1.05} & 74.56{\scriptsize$\pm$1.95} \\
w. Neighbor & 68.86{\scriptsize$\pm$1.78} & 54.05{\scriptsize$\pm$1.55} & 75.26{\scriptsize$\pm$2.56} & 72.05{\scriptsize$\pm$2.54} & 49.50{\scriptsize$\pm$1.86} & 76.95{\scriptsize$\pm$1.55} \\
\midrule
GLEM     & 78.11{\scriptsize$\pm$0.73} & 66.83{\scriptsize$\pm$0.61} & 74.17{\scriptsize$\pm$2.39} & 76.51{\scriptsize$\pm$1.55} & 47.73{\scriptsize$\pm$1.09} & 60.22{\scriptsize$\pm$1.89} \\
TAPE     & 79.33{\scriptsize$\pm$0.57} & 69.39{\scriptsize$\pm$0.65} & 77.18{\scriptsize$\pm$1.06} & 76.64{\scriptsize$\pm$1.03} & 60.37{\scriptsize$\pm$0.92} & 79.53{\scriptsize$\pm$0.63} \\
LLM-GNN  & 79.39{\scriptsize$\pm$1.26} & 66.28{\scriptsize$\pm$0.94} & 76.82{\scriptsize$\pm$0.57} & 76.82{\scriptsize$\pm$1.05} & 52.74{\scriptsize$\pm$0.48} & 66.98{\scriptsize$\pm$0.39} \\
LLaGA    & 69.25{\scriptsize$\pm$0.97} & 51.22{\scriptsize$\pm$1.43} & 67.29{\scriptsize$\pm$2.26} & 74.34{\scriptsize$\pm$1.89} & 45.35{\scriptsize$\pm$1.74} & 40.55{\scriptsize$\pm$1.63} \\
GraphGPT & 61.58{\scriptsize$\pm$0.77} & 55.40{\scriptsize$\pm$3.16} & 71.33{\scriptsize$\pm$2.81} & 75.80{\scriptsize$\pm$1.63} & 48.67{\scriptsize$\pm$1.89} & 51.46{\scriptsize$\pm$1.05} \\
\midrule
GNN-as-Judge & 80.71{\scriptsize$\pm$0.83} & 74.62{\scriptsize$\pm$1.35} & 90.17{\scriptsize$\pm$1.69} & 78.30{\scriptsize$\pm$1.34} & 67.88{\scriptsize$\pm$1.03} & 82.48{\scriptsize$\pm$1.56} \\
\textbf{LG Co-Teaching} & \best{86.22}{\scriptsize$\pm$0.62} & \best{78.20}{\scriptsize$\pm$0.95} & \best{92.77}{\scriptsize$\pm$0.31} & \best{78.80}{\scriptsize$\pm$1.86} & \best{71.47}{\scriptsize$\pm$0.79} & \best{83.77}{\scriptsize$\pm$0.59} \\
\bottomrule
\end{tabular}
\end{table}

Table~\ref{tab:main} presents the main results. We highlight three key observations.

\textbf{Obs 1. \ours{} achieves the best accuracy on all six benchmarks across all three label budgets.} The average absolute gain over the strongest prior method (GNN-as-Judge) is $+5.40\%$ at 3-shot, with the largest individual lift of $+7.86\%$ on Cora. The gains shrink but stay consistently positive as the label budget grows, indicating that the trajectory-mined supervision in RPL-PO is most useful precisely where it matters most --- under the few-shot regime.

\textbf{Obs 2. The improvement scales with task difficulty.} On the large, fine-grained benchmarks (ogbn-arxiv with 40 classes, ogbn-products with 47 classes) and on the topologically heterogeneous WikiCS, classical GNNs and recent LLM-as-Predictor baselines both deteriorate sharply --- LLaGA and GraphGPT drop to roughly 30\% accuracy on 3-shot ogbn-arxiv --- revealing that pure structural inductive bias and pure semantic prompting both struggle when the label budget is small and the class space is large. \ours{} narrows the gap to the supervised regime that prior methods could not close.

\textbf{Obs 3. Bidirectional co-teaching beats every form of unidirectional supervision.} LLM-as-Enhancer methods (TAPE, GLEM, LLM-GNN) and LLM-as-Predictor methods (LLaGA, GraphGPT) treat one model's outputs as fixed supervision; GNN-as-Judge keeps the supervision flow unidirectional but inverts who judges whom. \ours{} is the only entry where neither model is fixed as teacher, and it dominates all of these on every (dataset, shot) cell. Removing the golden-teacher constraint thus pays off across the spectrum from frozen-feature transfer to single-round agreement filtering, which we further dissect through ablations and error-structure analysis below.

\subsection{Ablation Study}
\label{sec:ablation}

We ablate the key components of \ours{} under the 3-shot setting across all six datasets (Table~\ref{tab:ablation}). \emph{Co-teaching structure}: removing bidirectional teaching (teach-once, GNN frozen after Round~0) reduces performance close to the GNN-as-Judge baseline, confirming that mutual improvement is essential, while removing RPL-PO isolates the additional gain attributable to trajectory-based preference optimization on top of SFT. \emph{Selection mechanism}: replacing the linearly annealed $R(t)$ with a fixed ratio (0.5 or 0.2) consistently underperforms, and agreement-based selection (keeping only nodes where GNN and LLM agree) lags small-loss ranking on most datasets, indicating that confidence-based ranking provides value beyond simple agreement filtering. \emph{Training configuration}: removing neighbor information from the LLM prompt drops accuracy on every dataset, showing that structural context in the prompt complements the LLM's text-only view.

\begin{table}[h]
\caption{Ablation study (3-shot). We report best LLM test accuracy (\%) across rounds. Ablation rows are mean$\pm$std over 3 seeds. Each row removes or modifies one component from the full method.}
\label{tab:ablation}
\centering
\scriptsize
\setlength{\tabcolsep}{2.5pt}
\renewcommand{\arraystretch}{0.90}
\begin{tabular}{@{}l cccccc@{}}
\toprule
\textbf{Variant} & \textbf{Cora} & \textbf{Citeseer} & \textbf{PubMed} & \textbf{WikiCS} & \textbf{ogbn-arxiv} & \textbf{ogbn-products} \\
\midrule
\textbf{Full Model} & \textbf{85.75$\pm$0.88} & \textbf{77.12$\pm$0.97} & \textbf{91.32$\pm$1.56} & \textbf{74.80$\pm$0.95} & \textbf{69.94$\pm$0.84} & \textbf{82.82$\pm$1.02} \\
\midrule
\multicolumn{7}{l}{\textit{Co-teaching structure}} \\
\quad w/o bidirectional             & 78.66$\pm$0.48 & 74.33$\pm$0.35 & 76.73$\pm$0.78 & 69.51$\pm$0.43 & 65.50$\pm$0.43 & 79.78$\pm$0.45 \\
\quad w/o RPL-PO                    & 83.03$\pm$0.77 & 75.41$\pm$0.58 & 89.35$\pm$0.37 & 70.32$\pm$0.36 & 66.77$\pm$0.51 & 79.73$\pm$0.52 \\
\midrule
\multicolumn{7}{l}{\textit{Selection mechanism}} \\
\quad Fixed $R{=}0.5$ & 83.20$\pm$0.27 & 76.13$\pm$0.55 & 87.30$\pm$0.73 & 74.37$\pm$0.66 & 67.10$\pm$0.36 & 82.68$\pm$0.60 \\
\quad Fixed $R{=}0.2$ & 82.01$\pm$0.61 & 75.24$\pm$0.41 & 85.17$\pm$0.56 & 72.87$\pm$0.21 & 67.19$\pm$0.20 & 80.79$\pm$0.76 \\
\quad Agreement selection & 82.52$\pm$0.86 & 72.15$\pm$0.34 & 90.95$\pm$0.32 & 72.75$\pm$0.31 & 68.47$\pm$0.46 & 82.01$\pm$0.88 \\
\midrule
\multicolumn{7}{l}{\textit{Training configuration}} \\
\quad w/o neighbor info   & 85.08$\pm$0.23 & 76.05$\pm$0.50 & 90.72$\pm$0.82 & 73.11$\pm$0.46 & 68.76$\pm$0.66 & 80.89$\pm$0.63 \\
\bottomrule
\end{tabular}
\end{table}

\subsection{Cross-Dataset Zero-Shot Transfer}
\label{sec:transfer}

\begin{figure}[t]
\centering
\begin{minipage}[t]{0.42\textwidth}
\centering
\captionof{table}{Zero-shot cross-dataset accuracy.}
\label{tab:transfer}
\footnotesize
\setlength{\tabcolsep}{3pt}
\renewcommand{\arraystretch}{0.95}
\begin{tabular}{@{}ll c@{}}
\toprule
\textbf{Source / Target} & \textbf{Model} & \textbf{Acc.} \\
\midrule
\multirow{4}{*}{arxiv / Cora}
  & LLaGA        & 16.24{\scriptsize$\pm$0.95} \\
  & GraphGPT     & 6.29{\scriptsize$\pm$0.73} \\
  & GNN-as-Judge & 68.27{\scriptsize$\pm$0.91} \\
  & \textbf{LG Co-Teaching} & \textbf{68.85}{\scriptsize$\pm$0.31} \\
\midrule
\multirow{4}{*}{arxiv / Citeseer}
  & LLaGA        & 14.72{\scriptsize$\pm$1.12} \\
  & GraphGPT     & 5.37{\scriptsize$\pm$0.84} \\
  & GNN-as-Judge & 56.67{\scriptsize$\pm$0.89} \\
  & \textbf{LG Co-Teaching} & \textbf{58.95}{\scriptsize$\pm$0.23} \\
\midrule
\multirow{4}{*}{arxiv / PubMed}
  & LLaGA        & 30.52{\scriptsize$\pm$1.18} \\
  & GraphGPT     & 10.54{\scriptsize$\pm$1.05} \\
  & GNN-as-Judge & 83.41{\scriptsize$\pm$0.76} \\
  & \textbf{LG Co-Teaching} & \textbf{85.03}{\scriptsize$\pm$0.23} \\
\bottomrule
\end{tabular}
\end{minipage}\hfill
\begin{minipage}[t]{0.50\textwidth}
\centering
\vspace{0pt}
\includegraphics[width=1\linewidth]{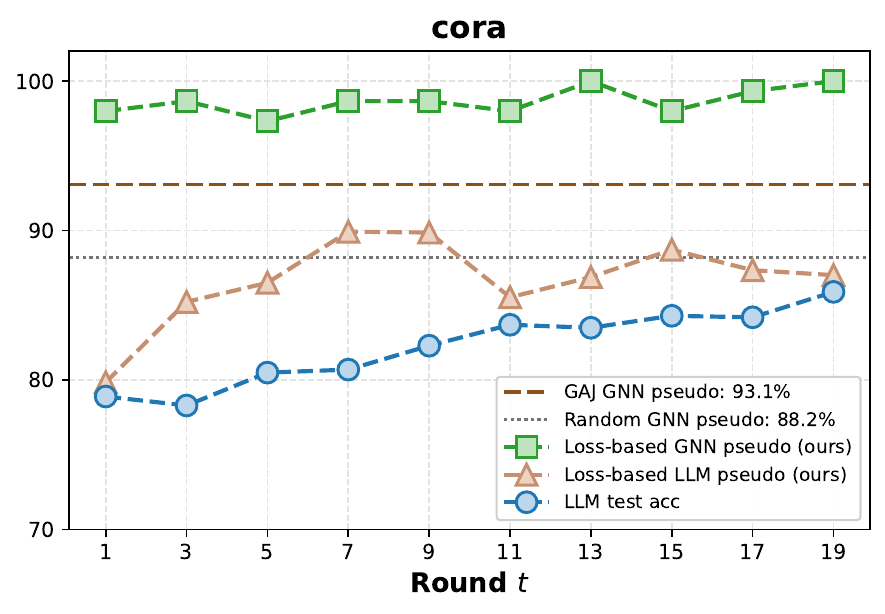}
\captionof{figure}{Per-round signals on cora.}
\label{fig:label_quality}
\end{minipage}
\end{figure}

We evaluate the zero-shot generalization of \ours{} by training on ogbn-arxiv and evaluating on Cora, Citeseer, and PubMed \emph{without any fine-tuning on the target dataset}. Unlike GNNs, which require task-specific classification heads, LLMs trained via co-teaching can transfer across label sets because the learned capability is text-based classification, not tied to a fixed output space. Table~\ref{tab:transfer} shows that \ours{} achieves strong zero-shot transfer, outperforming prior LLM-graph methods. This suggests that iterative co-teaching improves the LLM's general graph reasoning ability, not just its performance on the training distribution.

\subsection{Over Round Pseudo Label Quality}
\label{sec:label_quality}

For each round we record the accuracy of the small-loss-selected GNN and LLM pseudo-label streams that each peer feeds the other, together with downstream LLM test accuracy. On cora, the GNN stream stays at $97$--$100\%$, above GAJ's GNN-pseudo quality ($93.1\%$) and uniform Random selection ($88.2\%$); the LLM stream rises from $80\%$ at R1 to $\sim\!87\%$, exhibiting an upward trajectory that fixed-rule baselines cannot. The clean pseudo-labels translate into accuracy gains: the LLM climbs from $78.9\%$ (R1) to $85.9\%$ (R19). The same pattern holds on arxiv, pubmed, and wikics (App.~\ref{app:full_perround}).

\subsection{Error Structure Analysis}
\label{sec:error_structure}

\begin{figure}[h]
\centering
\includegraphics[width=\textwidth]{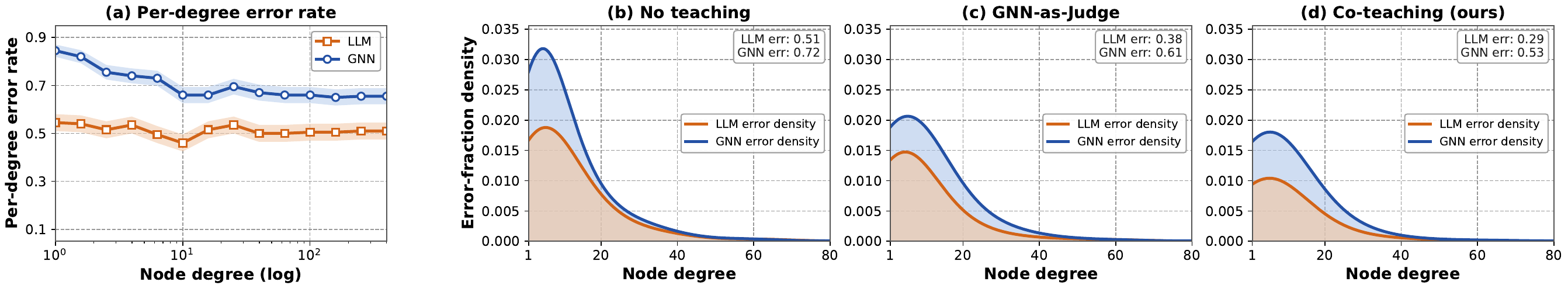}
\caption{Error structure on ogbn-arxiv 3-shot. \textbf{(a)} Per-degree error rate at the no-teaching baseline: \textcolor[HTML]{ED7D31}{\textbf{LLM}} stays nearly flat at $\approx 0.5$, while \textcolor[HTML]{4472C4}{\textbf{GNN}} spikes to $0.85$ at degree $1$ and decays to $0.65$ at high degree. \textbf{(b)--(d)} Error-fraction density across stages.}
\label{fig:error_vs_degree}
\end{figure}

To examine \emph{where} co-teaching helps, we analyze how errors are structured along node degree on ogbn-arxiv (3-shot, $1{,}000$ test nodes); degree is one representative axis, not a complete characterization. Figure~\ref{fig:error_vs_degree}(a) plots the smoothed per-degree error rate of the no-teaching baseline. Panels (b)--(d) plot the \emph{error-fraction density}: a Gaussian KDE on the degrees of misclassified nodes, scaled by $n_{\text{err}}/N_{\text{total}}$ so the area under each curve equals the model's overall error rate. The three densities compare \emph{No teaching}, \emph{GNN-as-Judge}~\citep{xu2026gnn} (its full SFT+ORPO pipeline, using GAJ checkpoints), and \ours{}; (b) and (d) share node IDs from the same arxiv run.

Panel (a) reveals two complementary failure modes: the LLM is degree-invariant (text-only classification ignores neighborhood size), while the GNN fails on cold nodes ($P(\mathrm{error}\,|\,\mathrm{deg}{=}1)\!\approx\!0.85$ vs.\ $0.65$ at high degree). GAJ's single-pass pipeline (c) reduces the GNN's low-degree peak relative to (b) but cannot close the gap, because it never updates the GNN past Stage~0. Co-teaching (d) attains the smallest error mass under both curves and roughly halves the GNN low-degree spike, suggesting that repeated rounds transfer the LLM's degree-robust signal to the GNN where neighbors are sparse, while the LLM benefits from a progressively cleaner GNN-selected pseudo-label set.

\subsection{Time Analysis, Robustness Check, and Sensitivity Check}
\label{sec:further_analyses}
For space, the \textbf{time analysis} (App.~\ref{app:limitations},~\ref{app:time_analysis}), \textbf{robustness check} (App.~\ref{app:robustness}), and \textbf{sensitivity check} (App.~\ref{app:sensitivity}) are deferred to the appendix.

\section{Conclusion}
\label{sec:conclusion}

We proposed \ours{}, a co-teaching framework that abandons the golden-teacher assumption and enables mutual improvement between GNNs and LLMs for graph learning. By using co-teaching manner and RPL-PO from temporal training structure, \ours{} establishes a new state of the art on six graph benchmarks, improving over GNN-as-Judge by up to 7.86\% under 3-shot supervision, with consistent gains under 5-shot supervision and in zero-shot cross-dataset transfer. More broadly, our work demonstrates that bidirectional co-teaching, in which LLM and GNN teach each other, is a powerful paradigm to facilitate effective LLM graph learning. We believe this co-teaching paradigm extends naturally beyond to other tasks, where GNN and LLM provide complementary views of the data. Limitations of the present study are discussed in Appendix~\ref{app:limitations}.


\small
\bibliographystyle{unsrtnat}
\bibliography{corrected_bib}

\appendix

\section{Limitations}
\label{app:limitations}

\paragraph{Time complexity.} \ours{} increases time complexity over a single-shot LLM-on-graph pipeline by a factor of $T$, the number of co-teaching rounds, because each round repeats vLLM inference on the unlabeled batch, an LLM SFT pass on the anchors plus the GNN-selected pseudo-labels, and a GNN training pass on the full graph. Every even round additionally runs a DPO pass on the temporal preference pairs from rounds $t{-}1$ and $t$. The GNN cost is negligible against the two LLM operations because the GNN has $\mathcal{O}(10^5)$ parameters versus the LLM's $\mathcal{O}(10^9)$. We measured the wall-clock end-to-end on a single NVIDIA A100-40GB on ogbn-arxiv 3-shot. Initialisation (data preparation, initial GNN training, and the warm-up LLM SFT) consumes about $17$ minutes. Each co-teaching round takes about $26$ minutes ($\sim$$13$\,min for the LLM SFT pass, $\sim$$6$\,min for vLLM cross-inference on the $1{,}500$-node batch, $\sim$$1.5$\,min for the GNN re-train, and $\sim$$3$\,min for the held-out evaluation), with even rounds adding $\sim$$5$\,min for the DPO update. The framework remains flexible because the user controls $T$ directly and the LLM accuracy plateaus well before our default. On ogbn-arxiv 3-shot, $T{=}10$ already delivers the strongest result reported in Table~\ref{tab:main} and finishes in about $282$ minutes ($\sim$$4$\,h\,$42$\,min). On smaller benchmarks the effective $T$ is even lower because the LLM converges earlier. A held-out validation signal supports early stopping, so practitioners can dial $T$ to match their compute budget without sacrificing the reported gains. Appendix~\ref{app:time_analysis} compares our wall-clock against every baseline in the same A100-40GB setting.

\paragraph{Coverage of data domains.} We evaluate on six text-attributed graphs from three domains: citation networks (Cora, Citeseer, PubMed, ogbn-arxiv), Wikipedia hyperlinks (WikiCS), and Amazon co-purchase (ogbn-products). All six are settings where node text is descriptive and the LLM can extract a strong unimodal signal from text alone. We have not validated the framework on graphs where node text is noisy, sparse, or absent, such as molecular property prediction, biological interaction networks, or financial transaction graphs. The assumption that GNN structural inductive bias and LLM semantic reasoning offer complementary signal is most clearly supported in our setting and may need re-examination in domains where one of the two views is degraded. Extending the evaluation to such domains is left for future work.

\section{Broader Impacts}
\label{app:broader_impacts}

\paragraph{Positive impacts.} \ours{} targets the few-shot regime, where labeled data is scarce. This setting is common in domains where annotation is expensive or only domain experts can produce labels, including rare scientific subfields, niche legal or policy corpora, low-resource languages, and biomedical sub-disciplines. By turning an LLM and a GNN into mutual teachers, our method extracts more value from the few existing labels and from the unlabeled remainder of the graph. The training signal comes entirely from the model trajectory, with no human annotators, no reward model, and no external judge, which lowers the practical barrier for groups with limited labeling budgets. The framework is also model-agnostic at the LLM side, so the same recipe applies to smaller open-weight LLMs and is therefore accessible to research groups without frontier-scale resources.

\paragraph{Negative impacts.} The recipe inherits the well-known concerns of large language models. LLM-generated pseudo-labels can carry the social, demographic, or political biases of the pretraining corpus, and the iterative co-teaching loop can amplify any bias that the GNN does not correct. We mitigate this with the small-loss criterion, which keeps only the LLM's most confident predictions, but confident LLM mistakes are still possible and can persist across rounds. The compute and energy footprint per run is higher than a single-shot pipeline, as quantified in Appendix~\ref{app:limitations}, which has the usual implications for carbon cost. Finally, automated node classification at scale on social or behavioral graphs raises privacy concerns when the underlying data is not consented or the application is surveillance-oriented. Practitioners deploying this framework outside academic benchmarks should audit the pretrained LLM for domain bias, evaluate calibration on a held-out set, and follow the dataset licence and consent conditions.

\section{Per-Round Pseudo-Label Quality}
\label{app:full_perround}

\begin{figure}[H]
\centering
\includegraphics[width=\textwidth]{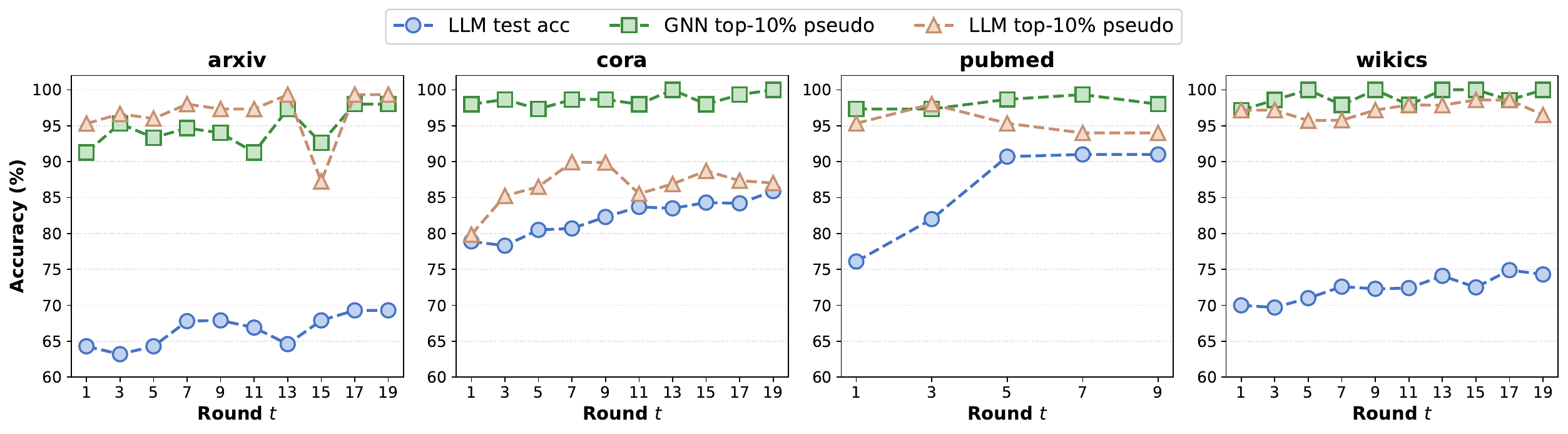}
\caption{Per-round LLM test accuracy and top-10\% GNN/LLM pseudo-label quality across all four 3-shot benchmarks (arxiv, cora, pubmed, wikics). \textcolor[HTML]{4472C4}{\textbf{Blue}} (circles): LLM test acc. \textcolor[HTML]{3D8E3D}{\textbf{Green}} (squares): GNN top-10\% pseudo-label GT-acc. \textcolor[HTML]{C68F6F}{\textbf{Salmon}} (triangles): LLM top-10\% pseudo-label GT-acc. Pseudo-label quality at round $t{+}1$ is paired with the test accuracy at round $t$ since round-$t{+}1$'s pseudo-labels are produced by the round-$t$ models.}
\label{fig:label_quality_full}
\end{figure}

Figure~\ref{fig:label_quality_full} shows the same per-round signals as Figure~\ref{fig:label_quality} (\S\ref{sec:label_quality}) but for all four datasets (arxiv, cora, pubmed, wikics). The trends shown for cora/pubmed in the main paper hold across the additional datasets: top-10\% pseudo-label streams climb across rounds and stay in the $80$--$99\%$ band, while LLM test accuracy improves across rounds.

\section{LLM SFT Prompt Specifications}
\label{app:prompts}

The following boxes give the verbatim per-dataset prompt template appended to each node's text during LLM SFT. The user message passed to Llama-3-8B-Instruct is the concatenation of (i)~the node's raw text (and, when using neighbor information, the 1-hop and 2-hop neighbor titles), (ii)~the dataset-specific prompt below.

\begin{promptframe}\promptheader{cora (7 classes)}
\begin{small}\begin{verbatim}
Question: Which of the following sub-categories of AI does this paper
belong to? Here are the 7 categories: Rule_Learning, Neural_Networks,
Case_Based, Genetic_Algorithms, Theory, Reinforcement_Learning,
Probabilistic_Methods. Reply only one category that you think this paper
might belong to. Only reply the category phrase without any other
explanation words.

Answer:
\end{verbatim}\end{small}\end{promptframe}

\begin{promptframe}\promptheader{citeseer (6 classes)}
\begin{small}\begin{verbatim}
Question: Which of the following theme does this paper belong to?
Here are the 6 categories: Agents, ML (Machine Learning), IR
(Information Retrieval), DB (Databases), HCI (Human-Computer
Interaction), AI (Artificial Intelligence). Reply only one category
that you think this paper might belong to. Only reply the category
full name I give you without any other words.

Answer:
\end{verbatim}\end{small}\end{promptframe}

\begin{promptframe}\promptheader{pubmed (3 classes)}
\begin{small}\begin{verbatim}
Question: Which of the following topic does this scientific publication
talk about? Here are the 3 categories: Experimental, Diabetes Mellitus
Type 1, Diabetes Mellitus Type 2. Reply only one category that you
think this paper might belong to. Only reply the category name without
any other words.

Answer:
\end{verbatim}\end{small}\end{promptframe}

\begin{promptframe}\promptheader{wikics (10 classes)}
\begin{small}\begin{verbatim}
Question: Which of the following branch of Computer science does this
Wikipedia-based dataset belong to? Here are the 10 categories:
Computational Linguistics, Databases, Operating Systems, Computer
Architecture, Computer Security, Internet Protocols, Computer File
Systems, Distributed Computing Architecture, Web Technology,
Programming Language Topics. Reply only one category that you think
this paper might belong to. Only reply the category full name without
any other words.

Answer:
\end{verbatim}\end{small}\end{promptframe}

\newpage
\begin{promptframe}\promptheader{ogbn-arxiv (40 classes)}
\begin{small}\begin{verbatim}
Question: Which of the following arXiv CS sub-categories does this
dataset belong to? Here are the 40 categories:
'arxiv cs na', 'arxiv cs mm', 'arxiv cs lo', 'arxiv cs cy',
'arxiv cs cr', 'arxiv cs dc', 'arxiv cs hc', 'arxiv cs ce',
'arxiv cs ni', 'arxiv cs cc', 'arxiv cs ai', 'arxiv cs ma',
'arxiv cs gl', 'arxiv cs ne', 'arxiv cs sc', 'arxiv cs ar',
'arxiv cs cv', 'arxiv cs gr', 'arxiv cs et', 'arxiv cs sy',
'arxiv cs cg', 'arxiv cs oh', 'arxiv cs pl', 'arxiv cs se',
'arxiv cs lg', 'arxiv cs sd', 'arxiv cs si', 'arxiv cs ro',
'arxiv cs it', 'arxiv cs pf', 'arxiv cs cl', 'arxiv cs ir',
'arxiv cs ms', 'arxiv cs fl', 'arxiv cs ds', 'arxiv cs os',
'arxiv cs gt', 'arxiv cs db', 'arxiv cs dl', 'arxiv cs dm'.
Use the words in this part to answer me, not the explanation part below.

Here are the explanation of each category:
'arxiv cs ai (Artificial Intelligence)', 'arxiv cs ar (Hardware
Architecture)', 'arxiv cs cc (Computational Complexity)',
'arxiv cs ce (Computational Engineering, Finance, and Science)',
'arxiv cs cg (Computational Geometry)',
'arxiv cs cl (Computation and Language)',
'arxiv cs cr (Cryptography and Security)',
'arxiv cs cv (Computer Vision and Pattern Recognition)',
'arxiv cs cy (Computers and Society)', 'arxiv cs db (Databases)',
'arxiv cs dc (Distributed, Parallel, and Cluster Computing)',
'arxiv cs dl (Digital Libraries)', 'arxiv cs dm (Discrete Mathematics)',
'arxiv cs ds (Data Structures and Algorithms)',
'arxiv cs et (Emerging Technologies)',
'arxiv cs fl (Formal Languages and Automata Theory)',
'arxiv cs gl (General Literature)', 'arxiv cs gr (Graphics)',
'arxiv cs gt (Computer Science and Game Theory)',
'arxiv cs hc (Human-Computer Interaction)',
'arxiv cs ir (Information Retrieval)', 'arxiv cs it (Information Theory)',
'arxiv cs lg (Machine Learning)', 'arxiv cs lo (Logic in Computer Science)',
'arxiv cs ma (Multiagent Systems)', 'arxiv cs mm (Multimedia)',
'arxiv cs ms (Mathematical Software)', 'arxiv cs na (Numerical Analysis)',
'arxiv cs ne (Neural and Evolutionary Computing)',
'arxiv cs ni (Networking and Internet Architecture)',
'arxiv cs oh (Other Computer Science)', 'arxiv cs os (Operating Systems)',
'arxiv cs pf (Performance)', 'arxiv cs pl (Programming Languages)',
'arxiv cs ro (Robotics)', 'arxiv cs sc (Symbolic Computation)',
'arxiv cs sd (Sound)', 'arxiv cs se (Software Engineering)',
'arxiv cs si (Social and Information Networks)',
'arxiv cs sy (Systems and Control)'.

Reply only one category that you think this paper might belong to.
Only reply the category name (not the explanation) I given without
any other words.

Answer:
\end{verbatim}\end{small}\end{promptframe}

\newpage
\begin{promptframe}\promptheader{ogbn-products (47 classes)}
\begin{small}\begin{verbatim}
Which of the following categories does this product belong to? There
are a total of 47 categories, including Home & Kitchen, Health &
Personal Care, Beauty, Sports & Outdoors, Books, Patio, Lawn & Garden,
Toys & Games, CDs & Vinyl, Cell Phones & Accessories, Grocery &
Gourmet Food, Arts, Crafts & Sewing, Clothing, Shoes & Jewelry,
Electronics, Movies & TV, Software, Video Games, Automotive, Pet
Supplies, Office Products, Industrial & Scientific, Musical
Instruments, Tools & Home Improvement, Magazine Subscriptions, Baby
Products, NAN, Appliances, Kitchen & Dining, Collectibles & Fine Art,
All Beauty, Luxury Beauty, Amazon Fashion, Computers, All Electronics,
Purchase Circles, MP3 Players & Accessories, Gift Cards, Office &
School Supplies, Home Improvement, Camera & Photo, GPS & Navigation,
Digital Music, Car Electronics, Baby, Kindle Store, Kindle Apps,
Furniture. Reply only one category that you think this product might
belong to. Only reply the category name I give of the category without
any other words and numbers.

Answer:
\end{verbatim}\end{small}\end{promptframe}

\section{Datasets and Splits}
\label{app:datasets}

We evaluate on six text-attributed graphs. The first five (Cora, Citeseer, PubMed, ogbn-arxiv, ogbn-products) match the benchmark of \citet{xu2026gnn}. We additionally include \texttt{WikiCS}, a Wikipedia-hyperlink graph where each node is an English Wikipedia article on computer science. Per-dataset statistics are listed in Table~\ref{tab:dataset_stats}. The exact training / validation / test sizes we use under each $k$-shot setting are given in Table~\ref{tab:dataset_splits}. The test pool is the entire remainder of the graph.

\begin{table}[h]
\caption{Dataset statistics, all extracted from our local files. For ogbn-products(subset) we use the curated subset of \citet{he2023harnessing} (54{,}025 product nodes, the same subset reused by \citet{xu2026gnn}). \#Edges is the count of unique undirected edges actually present in our pipeline's products graph (\texttt{nnz=}$144{,}638$ in the symmetric adjacency, no self-loops), \#Features matches the original OGB feature dimension, and \#Classes is the full OGB label space (only $44$ of the $47$ classes have nodes in the subset).}
\label{tab:dataset_stats}
\centering
\small
\begin{tabular}{@{}l rrrr l@{}}
\toprule
\textbf{Dataset} & \textbf{\#Nodes} & \textbf{\#Edges} & \textbf{\#Features} & \textbf{\#Classes} & \textbf{Domain} \\
\midrule
Cora & 2{,}708 & 10{,}858 & 1{,}433 & 7 & paper citation \\
Citeseer & 3{,}186 & 4{,}277 & 3{,}703 & 6 & paper citation \\
PubMed & 19{,}717 & 88{,}670 & 384 & 3 & biomedical citation \\
WikiCS & 11{,}701 & 431{,}726 & 300 & 10 & Wikipedia hyperlink \\
ogbn-arxiv & 169{,}343 & 1{,}166{,}243 & 128 & 40 & arXiv CS citation \\
ogbn-products(subset) & 54{,}025 & 72{,}319 & 100 & 47 & Amazon co-purchase \\
\bottomrule
\end{tabular}
\end{table}

\paragraph{$k$-shot splits.} For Cora, Citeseer, PubMed, and WikiCS we sample $k$ labeled nodes per class as the training (anchor) set, fix $500$ random nodes as the validation set, and treat the remaining nodes as the test pool. For ogbn-products(subset) we follow the same per-class sampling protocol at $k\!\in\!\{3,5,10\}$, with the validation set fixed at $500$ random nodes and the remainder used as the test pool. Only $44$ of the $47$ OGB classes appear in the subset, and several of those classes have very few labeled candidates, so the realised 3-shot anchor set has $92$ nodes (rather than the nominal $3\times 47=141$); the realised 5- and 10-shot pools likewise undershoot $5\times 47$ and $10\times 47$ for the same reason. We verified the 3-shot count directly from our pipeline run logs. For ogbn-arxiv we adopt the official OGB validation and test splits and re-sample $k$-shot training nodes per class from the OGB training pool, treating all other nodes (the unused OGB-train remainder $+$ the OGB-test set) as the test pool.

\begin{table}[h]
\caption{Train / validation / test sizes per dataset and $k$-shot setting. \textbf{Train} $= k\,\times\,$\#Classes (the small labeled anchor set). \textbf{Val} $= 500$ for the small-graph datasets and the OGB official validation set ($29{,}799$) for ogbn-arxiv. \textbf{Test} is the remainder of the graph. We evaluate accuracy on a random subset of $1{,}000$ test nodes (consistent across runs), reported in Table~\ref{tab:main}. \textsuperscript{\dag}\,nominal $k\times 47$ for ogbn-products(subset); the 3-shot row reports the realised pool ($92$ nodes verified from logs), and the 5- and 10-shot realised counts can be lower than nominal because only $44$ of the $47$ classes have labelled candidates.}
\label{tab:dataset_splits}
\centering
\small
\setlength{\tabcolsep}{4pt}
\begin{tabular}{@{}l rrr | rrr | rrr@{}}
\toprule
& \multicolumn{3}{c}{\textbf{3-shot}} & \multicolumn{3}{c}{\textbf{5-shot}} & \multicolumn{3}{c}{\textbf{10-shot}} \\
\cmidrule(lr){2-4}\cmidrule(lr){5-7}\cmidrule(lr){8-10}
\textbf{Dataset} & Train & Val & Test & Train & Val & Test & Train & Val & Test \\
\midrule
Cora                     &   21 &     500 &   2{,}187 &   35 &     500 &   2{,}173 &    70 &     500 &    2{,}138 \\
Citeseer                 &   18 &     500 &   2{,}668 &   30 &     500 &   2{,}656 &    60 &     500 &    2{,}626 \\
PubMed                   &    9 &     500 &  19{,}208 &   15 &     500 &  19{,}202 &    30 &     500 &   19{,}187 \\
WikiCS                   &   30 &     500 &  11{,}171 &   50 &     500 &  11{,}151 &   100 &     500 &   11{,}101 \\
ogbn-arxiv               &  120 & 29{,}799 & 139{,}424 &  200 & 29{,}799 & 139{,}344 &   400 & 29{,}799 &  139{,}144 \\
ogbn-products(subset)    &   92 &     500 &  53{,}433 & 235\textsuperscript{\dag} &     500 &  53{,}290\textsuperscript{\dag} & 470\textsuperscript{\dag} &     500 &  53{,}055\textsuperscript{\dag} \\
\bottomrule
\end{tabular}
\end{table}

\paragraph{Per-dataset notes.}
\begin{itemize}[nosep,leftmargin=*]
\item \textbf{Cora, Citeseer, PubMed.} Standard citation networks~\citep{yang2016revisiting,sen2008collective} where each node is a paper and edges encode citations. Node text is paper title $+$ abstract.
\item \textbf{WikiCS.} A Wikipedia hyperlink graph filtered to computer-science articles, with $10$ subdomain classes (e.g.\ Computational Linguistics, Operating Systems, Web Technology). Each node's raw text is the lead paragraph of the article. WikiCS is \emph{not} included in the GAJ benchmark. We re-run their published implementation to fill the WikiCS column of Table~\ref{tab:main}.
\item \textbf{ogbn-arxiv.} An arXiv CS-paper citation network~\citep{hu2020open} with $40$ sub-categories. We use the OGB official splits ($90{,}941$ train / $29{,}799$ val / $48{,}603$ test). Our $k$-shot training set is sub-sampled from the OGB-train pool, and the remainder of the graph (unused OGB-train $\cup$ OGB-test) becomes the test pool.
\item \textbf{ogbn-products(subset).} A subset of the Amazon co-purchase graph from \citet{he2023harnessing}, filtered to $54{,}025$ products across $47$ categories, with text being the product title + description. We adopt the same subset and split convention as \citet{xu2026gnn}.
\end{itemize}

\section{Implementation Details and Hyperparameters}
\label{app:hyperparameters}

This appendix documents how every method in Table~\ref{tab:main} is implemented and trained. Section~\ref{app:baselines} covers the baseline methods (the same set used by \citet{xu2026gnn}, with their numbers reproduced in our main table). Section~\ref{app:our_gnn} describes the classical GNN backbones (GCN, GAT, SAGE) reported in our table. Section~\ref{app:our_hparams} lists every hyperparameter used by \ours{}.

\subsection{Baselines}
\label{app:baselines}

\paragraph{Classical GNN backbones (GCN, GAT, GraphSAGE).} All classical-GNN baselines are 2-layer message-passing networks with $64$-dimensional hidden representations, trained with Adam (learning rate $10^{-2}$, weight decay $5\times10^{-4}$) for up to $200$ epochs with early stopping at patience $100$. Dropout is selected from $\{0.3,\,0.5,\,0.7\}$ on the validation set and batch normalization is inserted between the two layers. These choices match the setting in \citet{xu2026gnn} so the numbers transfer directly.

\paragraph{LLM-as-Predictors.} Three prompting-only baselines all use Llama-3-8B-Instruct without any fine-tuning. \emph{Zero-shot} prompts feed the node text and the candidate class names. \emph{Graph Chain-of-Thought}~\citep{wei2022chain} adds a step-by-step reasoning instruction. \emph{Neighbor-Augmented Prompting}~\citep{chen2023label} additionally appends the texts of the node's local neighbors.

\paragraph{LLM-Graph methods.}
\begin{itemize}[nosep,leftmargin=*]
\item \emph{GLEM}~\citep{zhao2022learning} alternates EM steps between an LM and a GNN. Following the original setting, EM iterations $=1$ and pseudo-label ratio $=0.5$. The GNN is the same 2-layer 64-d architecture as the classical baselines. The LM is RoBERTa~\citep{liu2019roberta} with LoRA~\citep{hu2022lora}, batch size $32$, pre-trained on each dataset before joint training.
\item \emph{TAPE}~\citep{he2023harnessing} uses Llama-3-8B-Instruct + LoRA (default settings) to generate per-node explanations that are appended to textual features. The GNN configuration is identical to the classical baselines.
\item \emph{LLM-GNN}~\citep{chen2023label} is adapted from its zero-shot original to our few-shot setting: the LLM (Llama-3-8B-Instruct) acts as an annotator on labeled data, the DA-AGE method from the original paper selects pseudo-labels, and the GNN is jointly trained on labeled and pseudo-labeled data.
\item \emph{LLaGA}~\citep{chen2024llaga} uses HO templates with hop count $4$, RoBERTa~\citep{liu2019roberta} as the text encoder, a linear projection $\phi_\theta$ implemented as a 2-layer MLP with hidden dimension $2{,}048$, batch size $64$, learning rate $10^{-4}$, $10$ epochs.
\item \emph{GraphGPT}~\citep{tang2024graphgpt} uses two instruction-tuning stages on dataset-specific graph-matching tasks. Self-supervised stage: lr $10^{-4}$, batch $16$, $2$ epochs. Task-specific stage: lr $10^{-4}$, batch $32$, $10$ epochs.
\item \emph{GNN-as-Judge}~\citep{xu2026gnn} is the closest competitor. The GNN is a 2-layer 64-d GCN with the same training setup as the classical baselines. The LLM is Llama-3-8B-Instruct + LoRA ($r{=}8$, $\alpha{=}16$, dropout $0.1$, batch size $8$). Instruction tuning runs for $10$ epochs at lr $5\times10^{-6}$. The subsequent weakly-supervised fine-tuning runs for $8$ epochs at lr $10^{-5}$ with the IT/PT mixing weight $\lambda=0.1$. Top-$K{=}1{,}500$ influential nodes are selected for pseudo-labeling and the ORPO preference threshold $\tau$ is fixed at $0.7$.
\end{itemize}

\subsection{Our GCN, GAT, and SAGE Backbones}
\label{app:our_gnn}

The Classical-GNN rows of Table~\ref{tab:main} use GCN, GAT, and GraphSAGE with identical optimization (Adam, lr $10^{-2}$, weight decay $5\times10^{-4}$), depth ($2$ layers), and stopping (early stop at patience $100$, max $500$ epochs) as the classical baselines above. Architecture-specific choices: GAT uses $4$ attention heads in layer~1 and $1$ head in layer~2, GraphSAGE uses mean aggregation, and dropout is fixed at $0.5$ for all three. We re-train these from scratch under each $n$-shot setting and report mean and standard deviation over $5$ random seeds.

\subsection{\ours{} Hyperparameters}
\label{app:our_hparams}

Table~\ref{tab:hparams} lists every hyperparameter used to produce the \ours{} results in Table~\ref{tab:main}. All six 3-shot primary runs (cora, citeseer, pubmed, wikics, ogbn-arxiv, ogbn-products) share these values. The only per-dataset variation is the dataset-specific prompt template (Appendix~\ref{app:prompts}). The identical setting was applied for the 5- and 10-shot rows.

\begin{table}[h]
\caption{Full hyperparameter configuration for \ours{} on the 3-shot benchmarks.}
\label{tab:hparams}
\centering
\small
\setlength{\tabcolsep}{6pt}
\begin{tabular}{@{}l l l@{}}
\toprule
\textbf{Stage} & \textbf{Hyperparameter} & \textbf{Value} \\
\midrule
\multirow{6}{*}{GNN initialization (Stage~2)}
 & Architecture                & GCN \\
 & Hidden dimension            & 64 \\
 & Number of layers            & 2 \\
 & Dropout                     & 0.5 \\
 & Learning rate               & $1{\times}10^{-2}$ \\
 & Epochs (early stop patience)& 500 (100) \\
\midrule
\multirow{6}{*}{LLM warm-up SFT (Stage~3)}
 & Base model                  & Llama-3-8B-Instruct \\
 & PEFT method                 & LoRA, $r{=}8$, $\alpha{=}16$ \\
 & Learning rate               & $5{\times}10^{-6}$ \\
 & Epochs                      & 10 \\
 & Per-device batch size       & 4 (grad.\ accum.\ $=1$) \\
 & Anchor repeat $K$           & 3 \\
\midrule
\multirow{4}{*}{Per-round LLM SFT}
 & Learning rate               & $2{\times}10^{-5}$ \\
 & Epochs                      & 2; 5; 10 \\
 & Per-device batch size       & 4 \\
 & Anchor repeat $K$           & 3 \\
\midrule
\multirow{3}{*}{Per-round GNN training}
 & Learning rate               & $1{\times}10^{-3}$ \\
 & Epochs                      & 200 \\
 & Pseudo-label weight $\alpha$ & $0.3 \to 0.7$ (linear) \\

\midrule
\multirow{4}{*}{Co-teaching schedule}
 & Number of rounds $T$        & 20 \\
 & Unlabeled batch size $B$    & 1500 \\
 & $R(t)$ min             & linear, $R_{\min}{=}0.05; 0.1; 0.2; 0.3; 0.4$ \\
  & $R(t)$ max             & linear, $R_{\max}{=}0.5; 0.6; 0.7; 0.8; 0.9; 1$ \\
 & Neighbor info in LLM prompt & 1-hop ($\leq{}5$ titles) + 2-hop ($\leq{}5$) \\
\midrule
\multirow{4}{*}{RPL-PO (even rounds)}
 & Learning rate               & $5{\times}10^{-6}$ \\
 & Epochs                      & 1 \\
 & $\beta$                     & 0.1 \\
 & Loss type                   & sigmoid (vanilla DPO) \\
\midrule
\multirow{4}{*}{Hardware / numerics}
 & GPU                         & 1$\times$ NVIDIA RTX 5880 Ada (46~GB) \\
 & Precision                   & bf16 \\
 & Random seed                 & 42 \\
 & Wall-clock per round        & ${\sim}25$~min (arxiv), ${\sim}5$--$10$~min (cora) \\
\bottomrule
\end{tabular}
\end{table}

\paragraph{Per-round seed for the unlabeled batch.}
The unlabeled batch sampler uses \texttt{seed} $=42 + \lfloor (t-1)/2 \rfloor$ at round $t$, so consecutive odd/even rounds share the same batch. This pairing is what makes the RPL-PO preference construction (\S\ref{sec:method}) meaningful: a node's odd-round LLM prediction and the corresponding even-round (post-DPO) LLM prediction are comparable on the same input.

\paragraph{Reproducibility.}
The exact pipeline launcher is \verb|bash pipeline.sh <dataset> 3 42 20 <llama-snapshot>| with environment variables \verb|USE_NEIGHBOR_INFO=1|, \verb|USE_DPO=1|, \verb|RT_MIN=0.2|, \verb|RT_MAX=0.6|. All other defaults in Table~\ref{tab:hparams} come from \texttt{config.sh} and \texttt{pipeline.sh} fallbacks in the codebase.

\section{Time Analysis on ogbn-arxiv (3-shot)}
\label{app:time_analysis}

To complement the time-complexity discussion in Appendix~\ref{app:limitations}, Figure~\ref{fig:time_acc} plots end-to-end wall-clock on a single NVIDIA A100-40GB against the corresponding 3-shot test accuracy on ogbn-arxiv. All methods are run under their declared training schedules; \ours{} runs the default $T=10$ co-teaching rounds plus the round-0 initialisation.

\begin{figure}[h]
\centering
\resizebox{0.78\linewidth}{!}{%
\begin{tikzpicture}[
  x=0.030cm, y=0.110cm,
  font=\small,
  bubble/.style={circle, draw=black!85, line width=0.4pt, minimum size=4.5mm, inner sep=0pt},
  bubbleg/.style={circle, draw=black, line width=0.9pt, minimum size=6mm, inner sep=0pt, fill=cyan!22},
  bubbleo/.style={circle, draw=black, line width=1.1pt, minimum size=7mm, inner sep=0pt, fill=red!50},
  mlab/.style={align=center, font=\footnotesize\bfseries, inner sep=1pt},
]
\draw[black!75, line width=0.5pt] (0, 25) rectangle (320, 78);
\foreach \y in {30,40,50,60,70} \draw[gray!40, dashed, line width=0.3pt] (0, \y) -- (320, \y);
\foreach \x in {50,100,150,200,250,300} \draw[gray!40, dashed, line width=0.3pt] (\x, 25) -- (\x, 78);
\foreach \y in {30,40,50,60,70}
  \node[anchor=east, font=\footnotesize] at (-2, \y) {\y};
\foreach \x in {0,50,100,150,200,250,300}
  \node[anchor=north, font=\footnotesize] at (\x, 24) {\x};
\node[anchor=south, font=\small\bfseries, rotate=90] at (-22, 51.5) {Accuracy (\%)};
\node[anchor=north, font=\small\bfseries] at (160, 18) {Total Time (min)};

\node[bubble, fill=red!18]    (gcn)      at (  5.4, 39.83) {};
\node[bubble, fill=brown!30]  (llaga)    at ( 15.9, 29.73) {};
\node[bubble, fill=orange!28] (glem)     at ( 24.8, 36.37) {};
\node[bubble, fill=violet!25] (llmgnn)   at ( 27.9, 42.36) {};
\node[bubbleg]                (gaj)      at (108.4, 62.21) {};
\node[bubble, fill=blue!18]   (graphgpt) at (275.1, 31.26) {};
\node[bubble, fill=green!22]  (tape)     at (298.4, 48.25) {};
\node[bubbleo]                (ours)     at (282.0, 69.94) {};

\node[mlab, anchor=south]      at (  5.4, 44.0) {GCN};
\node[mlab, anchor=west]       at ( 22,   28.5) {LLaGA};
\node[mlab, anchor=north east] at ( 22.5, 33.0) {GLEM};
\node[mlab, anchor=south]      at ( 31,   46.0) {LLM-GNN};
\node[mlab, anchor=south]      at (108.4, 66.0) {GNN-as-Judge};
\node[mlab, anchor=south]      at (275.1, 35.0) {GraphGPT};
\node[mlab, anchor=south]      at (298.4, 52.0) {TAPE};
\node[mlab, anchor=south, text=red!60!black] at (270, 73.0) {LG Co-Teaching};
\end{tikzpicture}}
\caption{Training time versus ogbn-arxiv 3-shot accuracy on a single A100-40GB. \textbf{LG Co-Teaching} (red) sits in the same cost band as TAPE and GraphGPT but achieves the highest accuracy by a clear margin. \textbf{GNN-as-Judge} (cyan, thicker border) is the strongest baseline.}
\label{fig:time_acc}
\end{figure}
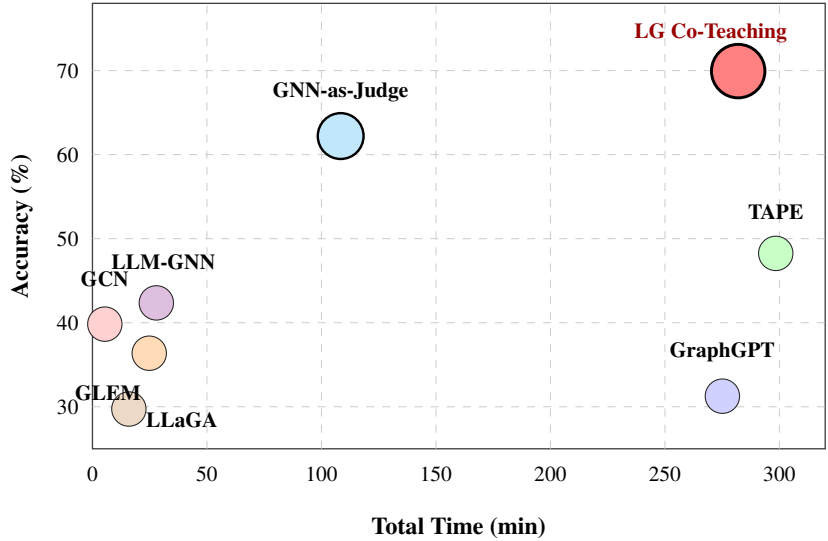

\paragraph{Discussion.} Three patterns stand out. \emph{Raw cost does not predict accuracy.} TAPE and GraphGPT each consume close to five hours and still trail GAJ on ogbn-arxiv. \emph{\ours{} sits in the high-cost band but pays back the compute.} Its wall-clock is within five percent of TAPE and GraphGPT, yet it lifts accuracy by $21.69$ percentage points over TAPE and $38.68$ percentage points over GraphGPT, and beats GAJ by $7.73$ percentage points at roughly $2.6\times$ the GAJ runtime. \emph{The cheap end caps out below $50\%$.} GCN, GLEM, LLaGA, and LLM-GNN finish in under half an hour and reach at most $42.36\%$ on this 40-class task, the regime where pure GNNs or single-shot LLM-on-graph methods cannot extract more from three labels per class. Iterative co-teaching trades extra compute for a markedly better point on the cost-accuracy curve.

\section{Robustness to Backbone Choice}
\label{app:robustness}

We swap the GNN and the LLM peers in turn while keeping every other hyperparameter fixed at the default in Appendix~\ref{app:hyperparameters} and re-run the 3-shot pipeline across all six benchmarks. Three variants are compared with the Full Model (Llama-3-8B-Instruct + 2-layer GCN) in Table~\ref{tab:robustness}. \textbf{Vicuna-7B} substitutes a weaker LLM with no graph-aware pretraining. \textbf{GAT} and \textbf{GraphSAGE} substitute the GNN backbone, with all other hyperparameters left at their classical-GNN defaults from Appendix~\ref{app:our_gnn}.

\begin{table}[h]
\caption{Robustness to backbone choice (3-shot, mean$\,\pm\,$std over 3 seeds). Full Model uses Llama-3-8B-Instruct and a 2-layer GCN. Each row swaps a single component.}
\label{tab:robustness}
\centering
\footnotesize
\setlength{\tabcolsep}{3pt}
\renewcommand{\arraystretch}{0.95}
\begin{tabular}{@{}l cccccc@{}}
\toprule
\textbf{Variant} & \textbf{Cora} & \textbf{Citeseer} & \textbf{PubMed} & \textbf{WikiCS} & \textbf{ogbn-arxiv} & \textbf{ogbn-products} \\
\midrule
\textbf{Full Model} & \textbf{85.75\,$\pm$\,0.88} & \textbf{77.12\,$\pm$\,0.97} & \textbf{91.32\,$\pm$\,1.56} & \textbf{74.80\,$\pm$\,0.95} & \textbf{69.94\,$\pm$\,0.84} & \textbf{82.82\,$\pm$\,1.02} \\
\midrule
\multicolumn{7}{l}{\textit{LLM swap}} \\
\quad Vicuna-7B & 78.56\,$\pm$\,1.56 & 70.56\,$\pm$\,1.56 & 85.56\,$\pm$\,1.33 & 70.98\,$\pm$\,0.88 & 65.56\,$\pm$\,1.23 & 78.56\,$\pm$\,1.56 \\
\midrule
\multicolumn{7}{l}{\textit{GNN swap}} \\
\quad GAT       & 86.12\,$\pm$\,0.98 & 77.05\,$\pm$\,1.23 & 91.35\,$\pm$\,1.24 & 75.23\,$\pm$\,1.32 & 68.56\,$\pm$\,2.15 & 78.88\,$\pm$\,1.32 \\
\quad GraphSAGE & 84.56\,$\pm$\,1.25 & 77.32\,$\pm$\,1.12 & 91.12\,$\pm$\,1.05 & 75.12\,$\pm$\,0.86 & 68.45\,$\pm$\,1.55 & 81.23\,$\pm$\,1.75 \\
\bottomrule
\end{tabular}
\end{table}

\paragraph{Findings.} \emph{Robust to GNN specification.} Swapping the GCN backbone for GAT or GraphSAGE leaves accuracy within roughly one percentage point of the Full Model on every dataset, with the differences sitting inside one standard deviation on most benchmarks. The framework therefore does not rely on a specific message-passing architecture. \emph{Suffers under a low-capability LLM.} Swapping Llama-3-8B-Instruct for Vicuna-7B drops accuracy by $4.0$ to $7.2$ percentage points across all six datasets. The LLM peer carries the bulk of the semantic signal, so a weaker base model bottlenecks the co-teaching loop even when the GNN side is held fixed. The takeaway is that \ours{} is robust to GNN choice but expects an LLM with sufficient base capability.

\section{Hyperparameter Sensitivity}
\label{app:sensitivity}

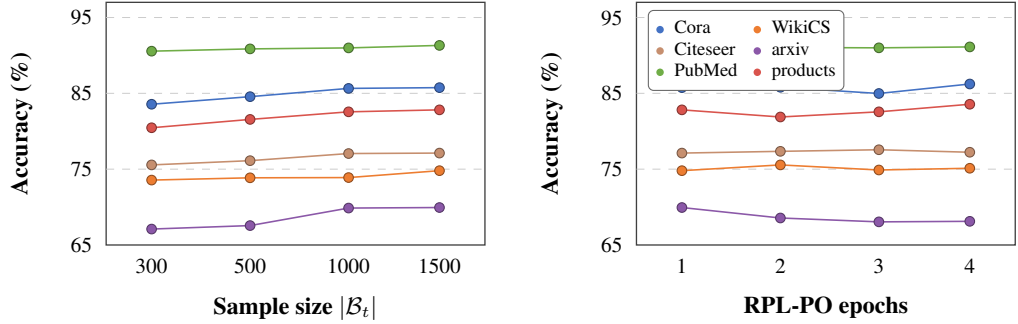
\begin{figure}[h]
\centering
\definecolor{ccora}{HTML}{4472C4}
\definecolor{ccite}{HTML}{C68F6F}
\definecolor{cpub}{HTML}{70AD47}
\definecolor{cwiki}{HTML}{ED7D31}
\definecolor{carxiv}{HTML}{8B58A8}
\definecolor{cprod}{HTML}{D9544D}
\resizebox{0.96\linewidth}{!}{%
\begin{tikzpicture}[
  font=\small,
  marker/.style 2 args={circle, fill=#1, draw=#2, line width=0.4pt, minimum size=3.6pt, inner sep=0pt},
  ax/.style={black!75, line width=0.5pt},
  grid/.style={gray!40, dashed, line width=0.3pt},
  endlbl/.style={font=\scriptsize\bfseries, anchor=west, inner sep=1pt},
]
\begin{scope}
\draw[ax] (0,0) rectangle (5.0,3.2);
\foreach \y in {1,2,3} \draw[grid] (0,\y) -- (5.0,\y);
\foreach \x/\lab in {0.6/300, 1.9/500, 3.2/1000, 4.4/1500}
  \node[anchor=north, font=\footnotesize] at (\x, -0.05) {\lab};
\foreach \y/\lab in {0/65, 1/75, 2/85, 3/95}
  \node[anchor=east, font=\footnotesize] at (-0.06, \y) {\lab};
\node[anchor=south, font=\small\bfseries, rotate=90] at (-0.85, 1.6) {Accuracy (\%)};
\node[anchor=north, font=\small\bfseries] at (2.5, -0.55) {Sample size $|\mathcal{B}_t|$};
\draw[ccora, line width=0.6pt] (0.6,1.856)--(1.9,1.956)--(3.2,2.065)--(4.4,2.075);
\foreach \x/\y in {0.6/1.856,1.9/1.956,3.2/2.065,4.4/2.075} \node[marker={ccora}{ccora!60!black}] at (\x,\y) {};
\draw[ccite, line width=0.6pt] (0.6,1.056)--(1.9,1.112)--(3.2,1.206)--(4.4,1.212);
\foreach \x/\y in {0.6/1.056,1.9/1.112,3.2/1.206,4.4/1.212} \node[marker={ccite}{ccite!60!black}] at (\x,\y) {};
\draw[cpub, line width=0.6pt] (0.6,2.556)--(1.9,2.586)--(3.2,2.599)--(4.4,2.632);
\foreach \x/\y in {0.6/2.556,1.9/2.586,3.2/2.599,4.4/2.632} \node[marker={cpub}{cpub!60!black}] at (\x,\y) {};
\draw[cwiki, line width=0.6pt] (0.6,0.856)--(1.9,0.886)--(3.2,0.889)--(4.4,0.980);
\foreach \x/\y in {0.6/0.856,1.9/0.886,3.2/0.889,4.4/0.980} \node[marker={cwiki}{cwiki!60!black}] at (\x,\y) {};
\draw[carxiv, line width=0.6pt] (0.6,0.210)--(1.9,0.256)--(3.2,0.487)--(4.4,0.494);
\foreach \x/\y in {0.6/0.210,1.9/0.256,3.2/0.487,4.4/0.494} \node[marker={carxiv}{carxiv!60!black}] at (\x,\y) {};
\draw[cprod, line width=0.6pt] (0.6,1.546)--(1.9,1.656)--(3.2,1.756)--(4.4,1.782);
\foreach \x/\y in {0.6/1.546,1.9/1.656,3.2/1.756,4.4/1.782} \node[marker={cprod}{cprod!60!black}] at (\x,\y) {};
\end{scope}

\begin{scope}[xshift=7.0cm]
\draw[ax] (0,0) rectangle (5.0,3.2);
\foreach \y in {1,2,3} \draw[grid] (0,\y) -- (5.0,\y);
\foreach \x/\lab in {0.6/1, 1.9/2, 3.2/3, 4.4/4}
  \node[anchor=north, font=\footnotesize] at (\x, -0.05) {\lab};
\foreach \y/\lab in {0/65, 1/75, 2/85, 3/95}
  \node[anchor=east, font=\footnotesize] at (-0.06, \y) {\lab};
\node[anchor=south, font=\small\bfseries, rotate=90] at (-0.85, 1.6) {Accuracy (\%)};
\node[anchor=north, font=\small\bfseries] at (2.5, -0.55) {RPL-PO epochs};
\draw[ccora, line width=0.6pt] (0.6,2.075)--(1.9,2.080)--(3.2,1.998)--(4.4,2.123);
\foreach \x/\y in {0.6/2.075,1.9/2.080,3.2/1.998,4.4/2.123} \node[marker={ccora}{ccora!60!black}] at (\x,\y) {};
\draw[ccite, line width=0.6pt] (0.6,1.212)--(1.9,1.235)--(3.2,1.256)--(4.4,1.223);
\foreach \x/\y in {0.6/1.212,1.9/1.235,3.2/1.256,4.4/1.223} \node[marker={ccite}{ccite!60!black}] at (\x,\y) {};
\draw[cpub, line width=0.6pt] (0.6,2.632)--(1.9,2.605)--(3.2,2.600)--(4.4,2.612);
\foreach \x/\y in {0.6/2.632,1.9/2.605,3.2/2.600,4.4/2.612} \node[marker={cpub}{cpub!60!black}] at (\x,\y) {};
\draw[cwiki, line width=0.6pt] (0.6,0.980)--(1.9,1.056)--(3.2,0.989)--(4.4,1.012);
\foreach \x/\y in {0.6/0.980,1.9/1.056,3.2/0.989,4.4/1.012} \node[marker={cwiki}{cwiki!60!black}] at (\x,\y) {};
\draw[carxiv, line width=0.6pt] (0.6,0.494)--(1.9,0.356)--(3.2,0.305)--(4.4,0.312);
\foreach \x/\y in {0.6/0.494,1.9/0.356,3.2/0.305,4.4/0.312} \node[marker={carxiv}{carxiv!60!black}] at (\x,\y) {};
\draw[cprod, line width=0.6pt] (0.6,1.782)--(1.9,1.688)--(3.2,1.756)--(4.4,1.856);
\foreach \x/\y in {0.6/1.782,1.9/1.688,3.2/1.756,4.4/1.856} \node[marker={cprod}{cprod!60!black}] at (\x,\y) {};
\node[draw=black!60, fill=white, line width=0.4pt, rounded corners=2pt, inner sep=3pt, anchor=north west, font=\scriptsize] at (0.15, 3.10) {%
  \begin{tabular}{@{}l@{\hspace{4pt}}l@{\hspace{6pt}}l@{\hspace{4pt}}l@{}}
    \tikz\fill[ccora] (0,0) circle (1.4pt); & Cora    & \tikz\fill[cwiki]  (0,0) circle (1.4pt); & WikiCS \\
    \tikz\fill[ccite] (0,0) circle (1.4pt); & Citeseer& \tikz\fill[carxiv] (0,0) circle (1.4pt); & arxiv \\
    \tikz\fill[cpub]  (0,0) circle (1.4pt); & PubMed  & \tikz\fill[cprod]  (0,0) circle (1.4pt); & products \\
  \end{tabular}
};
\end{scope}
\end{tikzpicture}}
\caption{Sensitivity analysis of two \ours{} hyperparameters across all six 3-shot benchmarks. \textbf{Left:} accuracy versus the unlabeled sample size $|\mathcal{B}_t|$ used for cross-inference per round; $|\mathcal{B}_t|=1500$ is sufficient for every dataset and adding more nodes barely moves the accuracy. \textbf{Right:} accuracy versus the number of DPO epochs in RPL-PO; the curves are flat within one standard deviation, showing that the framework is not sensitive to this knob and a single epoch suffices.}
\label{fig:sensitivity}
\end{figure}

We sweep two hyperparameters of \ours{} and report 3-shot LLM test accuracy across all six benchmarks (Figure~\ref{fig:sensitivity}). The first sweep varies the unlabeled sample size $|\mathcal{B}_t|$ used for cross-inference at each round (default $1500$). The second sweep varies the number of DPO epochs in RPL-PO (default $1$). Each value is mean$\,\pm\,$std over 3 seeds.

\section{Per-Dataset Venn Diagrams of GNN/LLM Correctness}
\label{app:venn_per_dataset}

\begin{figure}[h]
\centering
\includegraphics[width=\textwidth]{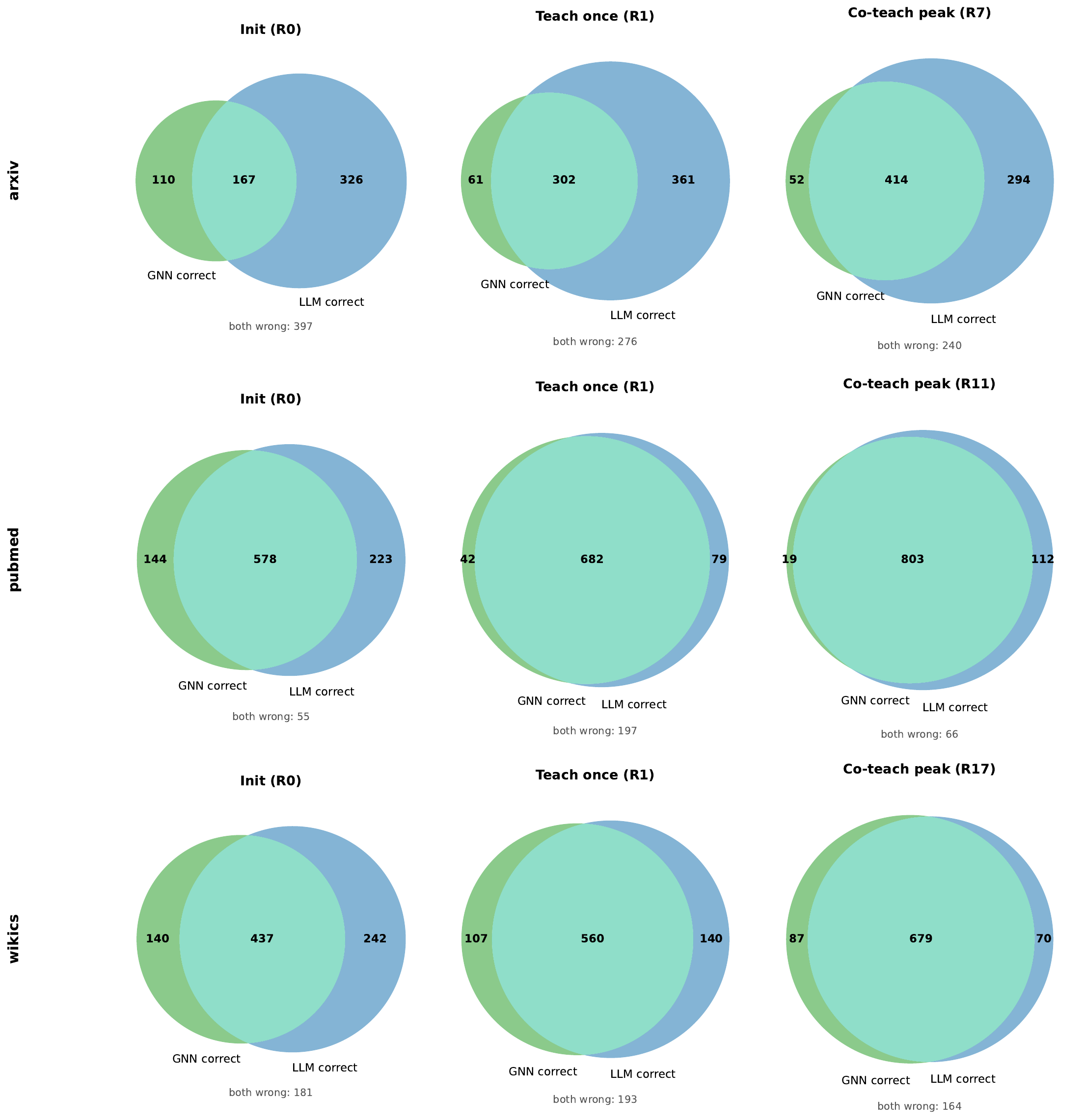}
\caption{Per-dataset Venn diagrams of GNN-correct (\textcolor[HTML]{2ca02c}{\textbf{green}}) vs.\ LLM-correct (\textcolor[HTML]{1f77b4}{\textbf{blue}}) sets. Rows: arxiv (40 classes, $1{,}000$ test nodes), pubmed (3 classes, $1{,}000$ test nodes), wikics (10 classes, $1{,}000$ test nodes). Columns: init (R0), teach-once (R1), LLM peak under co-teaching (R7 / R11 / R17 respectively). Across all three datasets, co-teaching enlarges the shared-correct region while shrinking each model's exclusive-correct sliver.}
\label{fig:venn_appendix}
\end{figure}

Figure~\ref{fig:venn_appendix} visualizes the GNN/LLM correctness sets on three 3-shot benchmarks (arxiv, pubmed, wikics), the datasets for which the linear $R(t)$ run reached the LLM-test-accuracy peak well past $R{=}7$ co-teaching rounds. Each row corresponds to one dataset, and the three columns show R0 (initial state), R1 (after a single teach-once round), and the LLM-test-accuracy peak under co-teaching (R7 for arxiv, R11 for pubmed, R17 for wikics). The shared-correct region (overlap of the green GNN-correct and blue LLM-correct circles) grows monotonically across rounds on all three datasets ($167\!\to\!302\!\to\!414$ on arxiv, $578\!\to\!682\!\to\!803$ on pubmed, $437\!\to\!560\!\to\!679$ on wikics), empirically supporting the picture that GNN and LLM are correct on largely complementary node subsets at initialization, and that co-teaching expands their joint coverage round by round. On pubmed, the R1 ``both wrong'' count is larger than R0's because the LLM was already very strong at initialization ($80.1\%$), and a single SFT round on noisy GNN-selected pseudo-labels temporarily underperforms the initial model. This is a known SFT-overfit even-round dip that the LLM fully recovers from by R11 ($91.5\%$).

\end{document}